\documentclass[manuscript,screen]{acmart}

\AtBeginDocument{%
  \providecommand\BibTeX{{%
    \normalfont B\kern-0.5em{\scshape i\kern-0.25em b}\kern-0.8em\TeX}}}

\setcopyright{acmcopyright}
\copyrightyear{2023}
\acmYear{2023}
\acmDOI{XXXXXXX.XXXXXXX}

\acmConference[Conference acronym 'XX]{Make sure to enter the correct
  conference title from your rights confirmation emai}{June 03--05,
  2018}{Woodstock, NY}
\acmPrice{15.00}
\acmISBN{978-1-4503-XXXX-X/18/06}

\usepackage{algorithmic}
\usepackage{algorithm}
\usepackage{multirow}
\begin{document}

\title{Mixed Graph Contrastive Network for Semi-Supervised Node Classification}

\author{Xihong Yang}
\email{yangxihong@nudt.edu.cn}
\affiliation{%
  \institution{National University of Defense Technology}
  \streetaddress{Deya road No.109}
  \city{ChangSha}
  \state{Hunan}
  \country{China}
  \postcode{410073}
}

\author{Yiqi Wang}
\affiliation{%
  \institution{National University of Defense Technology}
  \streetaddress{Deya road No.109}
  \city{ChangSha}
  \state{Hunan}
  \country{China}
  \postcode{410073}
}

\author{Yue Liu}
\affiliation{%
  \institution{National University of Defense Technology}
  \streetaddress{Deya road No.109}
  \city{ChangSha}
  \state{Hunan}
  \country{China}
  \postcode{410073}
}

\author{Yi Wen}
\affiliation{%
  \institution{National University of Defense Technology}
  \streetaddress{Deya road No.109}
  \city{ChangSha}
  \state{Hunan}
  \country{China}
  \postcode{410073}
}

\author{Lingyuan Meng}
\affiliation{%
  \institution{National University of Defense Technology}
  \streetaddress{Deya road No.109}
  \city{ChangSha}
  \state{Hunan}
  \country{China}
  \postcode{410073}
}

\author{Sihang Zhou}
\affiliation{%
  \institution{National University of Defense Technology}
  \streetaddress{Deya road No.109}
  \city{ChangSha}
  \state{Hunan}
  \country{China}
  \postcode{410073}
}

\author{Xinwang Liu}
\authornote{Corresponding Author}
\email{xinwangliu@nudt.edu.cn}
\author{En Zhu}
\authornotemark[1]
\affiliation{%
  \institution{National University of Defense Technology}
  \streetaddress{Deya road No.109}
  \city{ChangSha}
  \state{Hunan}
  \country{China}
  \postcode{410073}
}


\renewcommand{\shortauthors}{Yang, et al.}

\begin{abstract}
  Graph Neural Networks (GNNs) have achieved promising performance in semi-supervised node classification in recent years. However, the problem of insufficient supervision, together with representation collapse, largely limits the performance of the GNNs in this field. To alleviate the collapse of node representations in semi-supervised scenario, we propose a novel graph contrastive learning method, termed \textbf{M}ixed \textbf{G}raph \textbf{C}ontrastive \textbf{N}etwork (MGCN). In our method, we improve the discriminative capability of the latent embeddings by an interpolation-based augmentation strategy and a correlation reduction mechanism. Specifically, we first conduct the interpolation-based augmentation in the latent space and then force the prediction model to change linearly between samples. Second, we enable the learned network to tell apart samples across two interpolation-perturbed views through forcing the correlation matrix across views to approximate an identity matrix. By combining the two settings, we extract rich supervision information from both the abundant unlabeled nodes and the rare yet valuable labeled nodes for discriminative representation learning. Extensive experimental results on six datasets demonstrate the effectiveness and the generality of MGCN compared to the existing state-of-the-art methods. The code of MGCN is available at https://github.com/xihongyang1999/MGCN on Github. 
\end{abstract}




\maketitle

\section{Introduction}

In recent years, machine learning has developed rapidly and achieved remarkable performance in many fields, image classification \cite{tan1,tan2,tan3,he1,he2}, recommendation \cite{chen1,chen2,chen3,yin,zhu1}, and clustering \cite{wan1,wan2,wan3,DealMVC,qun,suyuan_AAAI,suyuan_TNNLS}. With the strong representation learning capacity, graph learning methods \cite{mo1,mo2,mo3,zhu2} have become a hot research spot in many fields, including the graph clustering \cite{CCGC,HSAN,Dink_net,CONVERT,liumeng,lu}, collaborative filtering \cite{zhao1,zhao2}, molecular graph \cite{zaixi1,zaixi2,xia1,xia2}, and so on. Semi-supervised node classification, which aims to classify nodes in the graph with limited labels, is a crucial yet challenging graph learning task. Thanks to the powerful feature extraction capability, Graph Convolutional Network (GCN) \cite{GCN} has recently achieved promising performance in this scenario. As a result, it has attracted considerable attention in this field, and many methods \cite{GAT, PPNP, JKNet, GPRGNN} have been proposed.

\begin{figure}[!t]
\centering
\small
\begin{minipage}{0.23\linewidth}
\centerline{\includegraphics[width=\textwidth]{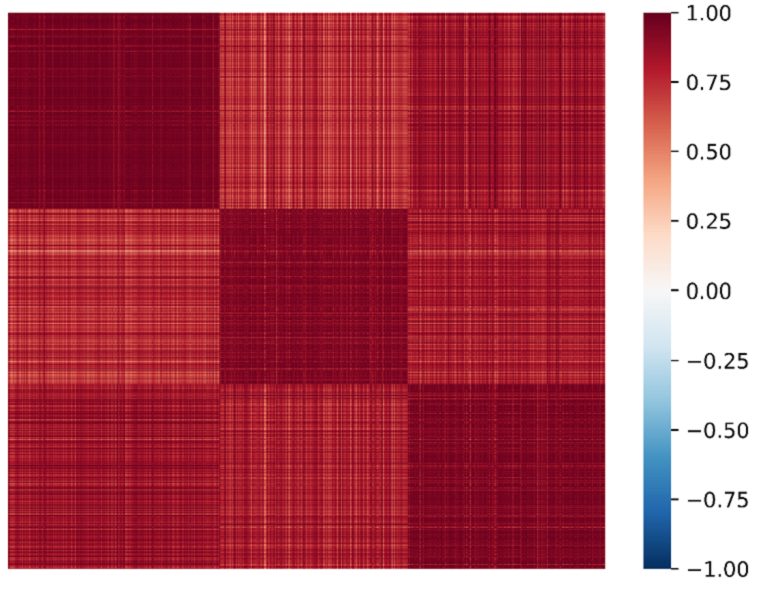}}
\centerline{(a) GCN}
\end{minipage}
\begin{minipage}{0.23\linewidth}
\centerline{\includegraphics[width=\textwidth]{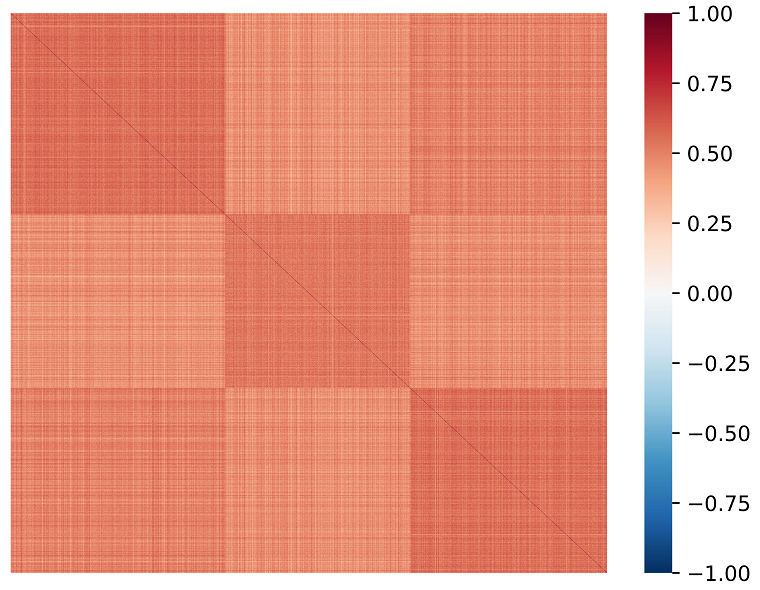}}
\centerline{(b) MixupForGraph}
\end{minipage}
\begin{minipage}{0.23\linewidth}
\centerline{\includegraphics[width=\textwidth]{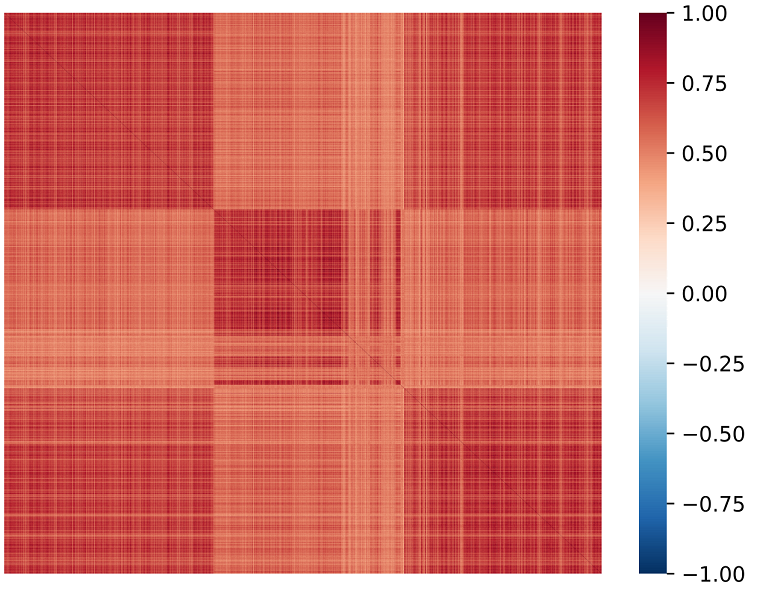}}
\centerline{(c) MVGRL}
\end{minipage}
\begin{minipage}{0.23\linewidth}
\centerline{\includegraphics[width=\textwidth]{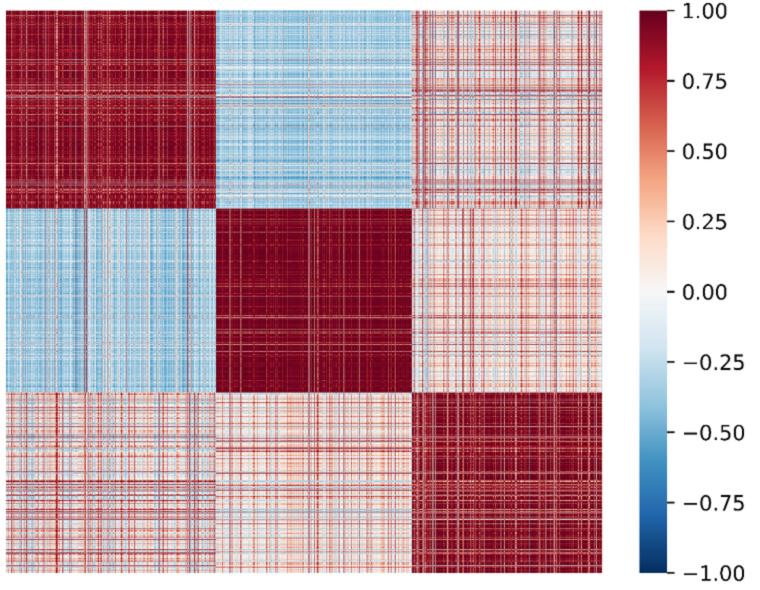}}
\centerline{(d) Ours}
\end{minipage}
\caption{{Visualization of cosine similarity matrices of the output embeddings of GCN \cite{GCN}, MixupForGraph \cite{GRAPH_MIXUP_1}, MVGRL\cite{MVGRL} and our proposed method on the ACM dataset. The sample order is rearranged to make samples from the same cluster beside each other. The higher value (\textcolor{red}{red}) indicates that embeddings are more similar, thus easy leading to representation collapsing. The lower value (\textcolor{blue}{blue}) denotes that the embeddings are less similar.}}
\label{MOTIVATION}  
\end{figure}


Although preferable performances have been achieved by the existing algorithms, in the semi-supervised node classification task, insufficient supervision has largely aggravated the problem of representation collapse in graph learning, leading to indiscriminative representation across classes. To solve the problem, a commonly used strategy is to path the supervision information from the labeled data to the unlabeled data according to the linkages within the adjacent matrix as guidance for network training \cite{GCN, GAT, JKNet, Graphsage,yq1,yq2,yq3}. Moreover, in MixupForGraph \cite{GRAPH_MIXUP_1}, a graph mixup operation is designed to enhance the robustness and discriminative capability of the aggregated sample embedding over the labeled samples. Since the embedding of the labeled samples has integrated information of both the labeled sample and its unlabeled neighbors while pushing the predictions to their corresponding ground truth, the information of the unlabeled samples are also integrated for network training in a form of implicit regularization. 

Though valuable information is introduced, the performance of these methods could be significantly influenced by the inaccurate connections within the data. Recently, to alleviate the adverse influence of the inaccurate connections, MVGRL\cite{MVGRL} introduces contrastive learning as an auxiliary task for discriminative information exploitation. In this method, the authors design an InfoMax loss to maximize the cross-view mutual information between the node and the global summary of the graph.

Although large improvement has been made, the current data augmentation and loss function setting of MVGRL fails to exploit abundant intuitive information within the unlabeled data thus limiting its classification performance.
This phenomenon can be witnessed in the cosine similarity matrix of latent representation illustration in Fig. \ref{MOTIVATION}. As we can see, although the categorical information is revealed by the learned representations to different extent, more discriminative information is needed for further performance enhancement.

To solve this issue, we propose a novel graph contrastive semi-supervised learning method termed \textbf{M}ixed \textbf{G}raph \textbf{C}ontrastive \textbf{N}etwork (MGCN), which improves the discriminative capability of node embedding by conducting the interpolation-based augmentation and improving the cross-view consistency of the latent representation among samples. To be specific, we first adopt the interpolation-based strategy to conduct data augmentation in the latent space and then force the prediction model to change linearly between samples as done in the field of image recognition \cite{ICT}. After that, by forcing the correlation matrix across two interpolation-perturbed views to approximate an identical matrix, we guide our network to be able to recognize whether two perturbed samples are the same samples or not.
In this manner, the sample representations would be more discriminative, thus alleviating the collapsed representations. This could be clearly seen in Fig. \ref{MOTIVATION} (d) that the similarity matrix generated by our method can obviously reveal the hidden distribution structure better than the compared methods. The key contributions of this paper are listed as follows:

\begin{itemize}

\item We observe the representation collapse problem under the semi-supervised scene, and propose a novel graph contrastive learning method to solve this issue.

\item An interpolation-based augmentation strategy and a correlation reduction mechanism are designed to improve discriminative capability of representations. 


\item Extensive experimental results on six datasets demonstrate the superiority of our method against the compared state-of-the-art method. The ablation study and module transferring experiments demonstrate the effectiveness and the generality of our proposed modules.

\end{itemize}

\section{Related Work}\label{relatedWork}

\subsection{Semi-supervised Node Classification}
Semi-supervised node classification \cite{wang_1,huang1,huang2,huang3} aims to classify nodes in the graph with few human annotations. Recently, Graph Neural Networks (GNNS) have achieved promising performance for their strong representation capability of graph-structured data. The pioneer GCN-Cheby \cite{GCN_CHEBY} generalizes CNN \cite{CNN} to graphs in the spectral domain by proposing the Chebyshev polynomials graph filter. Following GCN-Chey, GCN \cite{GCN} reveals the underlying graph structure by feature transformation and aggregation operations in the spatial domain. 
After that, GraphSage \cite{Graphsage} generates embeddings by sampling and aggregating features from the node neighborhoods. GAT\cite{GAT} proposed graph attention networks on graph-structured data to improve the performance. JK-Net\cite{JKNet} flexibly leverages different neighborhood ranges to enable better structure-aware representation. In addition, SGC\cite{SGC} simplifies GCN by removing feature transformation between consecutive layers. Furthermore, Geom-GCN\cite{GEOM-GCN} proposes a geometric aggregation scheme to overcome the issue of neighborhood node structural information loss. Different from them, PPNP/APPNP \cite{PPNP} separates the feature transformation from aggregation operation and enhances the aggregation operation with PageRank \cite{PAGERANK}. More recently, following PPNP/APPNP, GPRGNN\cite{GPRGNN} jointly optimizes sample feature and topological information by learning the aggregation weights adaptively. 

In our proposed method, we adopt GPRGNN \cite{GPRGNN} as our backbone and further improve its discriminative capability by conducting the interpolation-based aug and improving the cross-view consistency of the latent representation.

\subsection{Representation Collapse}
Contrastive learning methods \cite{SIMCLR,BYOL, IDCRN,DCRN} have achieved promising performance on images in recent years. Motivated by their success, contrastive learning strategies have been increasingly adopted to the graph data \cite{DGI,MVGRL,GRACE,SCGC,GCC-LDA}.

The pioneer DGI \cite{DGI} is proposed to learn node embedding by maximizing the mutual information between the local and global fields of the graph. GMI\cite{GMI} and HDMI\cite{HDMI} improve DGI by regarding edges and node attributes, respectively, to alleviate collapse representation.  Besides, MVGRL \cite{MVGRL} and InfoGraph \cite{InfoGraph} demonstrate the effectiveness of maximizing the mutual information to learn graph-level representations in the graph classification task. Subsequently, GraphCL \cite{GraphCL} and GRACE \cite{GRACE} first generate two augmented views and then learn node embeddings by pulling together the same node in two augmented views while pushing away different nodes. However, representation collapse is a common problem that, without the adequate guidance of human annotations, the model tends to embed all samples to the same representation.

In order to alleviate representation collapse, BGRL \cite{GRAPH_BYOL} is proposed to learn node embeddings by two separate GCN encoders. Specifically, the online encoder is trained to pull together the same node from two views while the target encoder is updated by an exponential moving average of online encoder. More recently, G-BT \cite{GBT} is proposed to avoid representation collapse by reducing the redundancy of features. MGCN implicitly achieves the redundancy-reduction principle through an interpolation-based correlation reduction mechanism in the sample level, described in section 3.3 to solve the representation collapse issue in the semi-supervised node classification task.

\subsection{Interpolation-based Augmentation}
Mixup\cite{mixuP,ICT} is an effective data augmentation strategy for image classification \cite{xihong}. It generates synthetic samples by linearly interpolating random image pairs and their labels as follows:  

\begin{equation}\label{mixup_}
  \begin{aligned}
    \lambda & \sim Beta(\alpha, \beta), \\
    {\lambda}^{\prime}& = max(\lambda, 1-\lambda), \\
    x^{\prime}&={\lambda}^{\prime}{x_1}+(1-{\lambda}^{\prime}){x_2},\\
    y^{\prime}&={\lambda}^{\prime}{y_1}+(1-{\lambda}^{\prime}){y_2},
  \end{aligned}
\end{equation}
where $\alpha$ and $\beta$ are the hyper-parameters of Beta distribution. Besides, $\lambda \in [0,1]$ denotes the interpolation rate. Actually, Mixup incorporates the prior knowledge that interpolations of input samples should lead to interpolations of the associated targets \cite{mixuP}. In this manner, it extends the training distribution by constructing virtual training samples across all classes, thus improving the image classification performance \cite{manifold,ICT}.

However, it is challenging to extend Mixup methods to the graph data, which contains many irregular connections. To solve this problem, GraphMixup \cite{graphmixup} designs feature and edge Mixup mechanisms to improve the performance of class-imbalanced node classification. Besides, MixupForGraph \cite{GRAPH_MIXUP_1} proposed the two-branch graph convolution to mix the receptive field sub-graphs for the paired nodes. Moreover, GraphMix \cite{graphmix} trains a fully-connected network(FCN) jointly with the graph neural network(GNN). The interpolated strategy in GraphMix is implemented with the node features ($\textbf{X}$) in FCN. Although the GNN could share parameters with FCN, the topology information still can’t be fully exploited. Different from the previous methods, we propose a simple interpolation fashion. Specially, we interpolate the embeddings and associated labels directly, which can simultaneously learn topology and feature information.

\begin{figure*}
\centering
\includegraphics[scale=0.6]{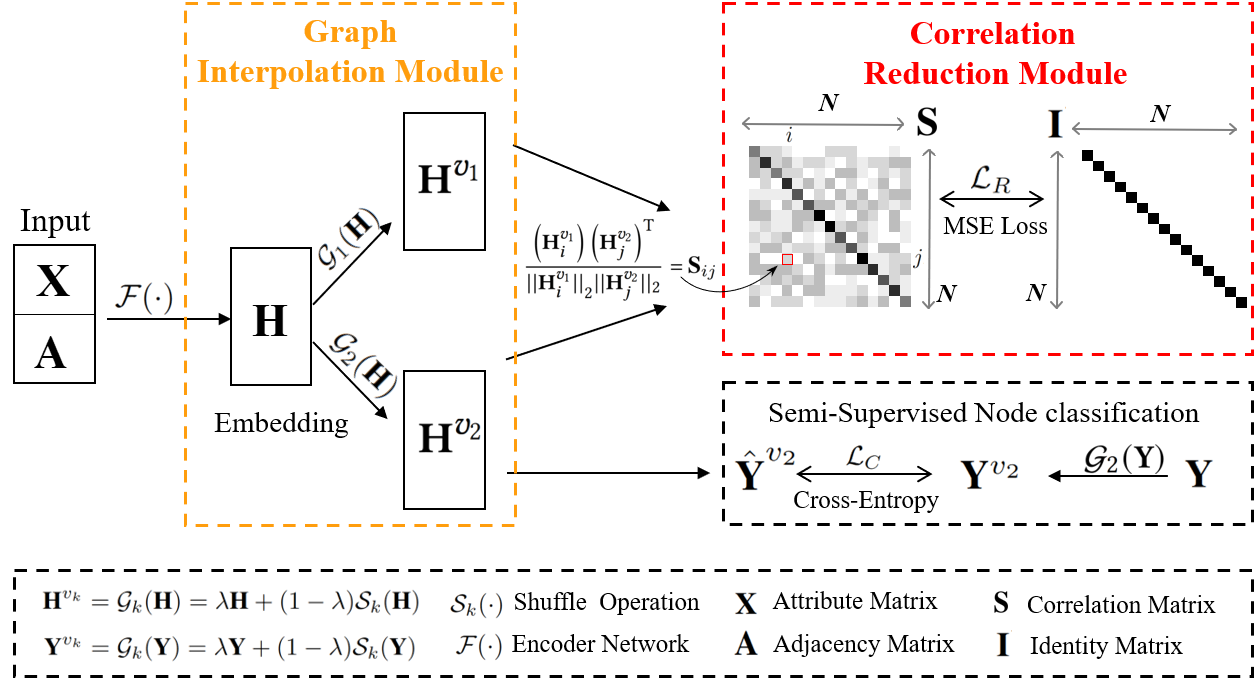} 
\vspace{5pt}
\caption{Illustration of \textbf{M}ixed \textbf{G}raph \textbf{C}ontrastive \textbf{N}etwork (MGCN). In the Graph Interpolation Module, with the generated embedding $\mathbf{H}$, we first adopt the interpolation-based strategy to conduct data augmentation in the latent space and then by guiding $\mathbf{H}^{v_2}$ to approximate the prediction $\mathbf{Y}^{v_2}$, we force the prediction model to change linearly between samples. Afterward, by guiding the cross-view correlation matrix to approximate the identity matrix, we enable the learned network to tell apart samples across two interpolation-perturbed views. In this manner, our network would be guided to learn the more discriminative embedding, thus alleviating representation collapse. In our model, the interpolation rate $\lambda$ is set as $0.95$ to make sure that $\mathbf{H}^{v_k}$ is a perturbation of $\mathbf{H}$.
}
\label{OVERRALL_FIGURE}  
\end{figure*}

\section{Method}

In this section, we proposed a novel graph contrastive learning method, termed \textbf{M}ixed \textbf{G}raph \textbf{C}ontrastive \textbf{N}etwork (MGCN), to improve the latent feature's discriminative capability and alleviate the collapsed representation. As shown in Fig.\ref{OVERRALL_FIGURE}, our proposed method mainly contains two modules, i.e., the graph interpolation module and correlation reduction module. In the following subsections, we first define the main notations and the problem. Then we detail the two main modules and loss function of MGCN.

\begin{table}[h]
\centering
\caption{Notation summary.}
\begin{tabular}{cc}
\hline
\textbf{Notations} & \textbf{Meaning} \\ \hline
$\textbf{X} \in \mathbb{R}^{N\times D}$                  & \textbf{The Attribute Matrix}                \\
$\textbf{A} \in \mathbb{R}^{N\times N}$                  & \textbf{The Adjacency Matrix}                \\
$\textbf{D} \in \mathbb{R}^{N\times N}$                  & \textbf{The Degree Matrix}                \\
$\textbf{I} \in \mathbb{R}^{N\times N}$                  & \textbf{The Identity Matrix}                \\
$\textbf{H} \in \mathbb{R}^{N\times D}$                  & \textbf{The Node Embeddings}                \\
$ \textbf{S} \in \mathbb{R}^{N\times N}$                  & \textbf{The Cross-view Sample Correlation Matrix}                \\
$\hat{\textbf{Y}} \in \mathbb{R}^{N\times C}$                  & \textbf{The Prediction Distribution}                \\
$\textbf{Y} \in \mathbb{R}^{N\times C}$                 & \textbf{The Label Distribution}                \\ \hline
\end{tabular}
\label{NOTATION_TABLE} 
\end{table}

\subsection{Notations and Problem Definition}
To an undirected graph $\left \{\mathcal{V}, \mathcal{E} \right \}$ with $K$ classes of nodes, the node set and the edge set are denoted as $\mathcal{V}=\{v_1, v_2, \dots, v_N\}$ and $\mathcal{E}$, respectively. The graph contains an attribute matrix $\textbf{X} \in \mathbb{R}^{N\times D}$ and an adjacency matrix $\textbf{A}=(a_{ij})_{N\times N}$, where $a_{ij}=1$ if $(v_i,v_j)\in \mathcal{E}$, otherwise $a_{ij}=0$. The degree matrix is denoted as $\textbf{D}=diag(d_1, \dots ,d_N) \in \mathbb{R}^{N\times N}$ and $d_i=\sum_{(v_i,v_j)\in \mathcal{E}}a_{ij}$. The normalized adjacency matrix $\widetilde{\textbf{A}} \in \mathbb{R}^{N\times N}$ could be calculated through calculating $\textbf{D}^{-\frac{1}{2}}(\textbf{A}+\textbf{I})\textbf{D}^{-\frac{1}{2}}$, where $\textbf{I} \in \mathbb{R}^{N \times N}$ is an identity matrix. Besides, $||\cdot||_2$ denotes the ${\ell _2}$-norm. In this paper, our target is to embed the nodes into the latent space and classify them in a semi-supervised manner. The notations are summarized in Table \ref{NOTATION_TABLE}.

\subsection{Graph Interpolation Module}

Recent works \cite{mixuP} demonstrate that Mixup is an effective data augmentation for images to improve the discriminative capability of samples by achieving larger margin-decision boundaries. Different from images, the nodes in the graph are irregularly connected. Thus, the interpolation for the graph data is still an open question \cite{graphmix,graphmixup}. 

To overcome this issue, we propose a simple yet effective interpolation method on graph data as shown in the orange box in Fig.\ref{OVERRALL_FIGURE}. Specifically, we first encode the nodes into the latent space through Eq. \eqref{GPRGNN_ENCODING}.
\begin{equation} 
{\textbf{H}} = \mathcal{F}(\textbf{X}, \textbf{A}). 
\label{GPRGNN_ENCODING}
\end{equation}
Here, $\mathcal{F}(\cdot)$ denotes the encoder of our feature extraction framework. In our paper, we take the encoder of GPRGNN \cite{GPRGNN}, which learns node embeddings from node features and topological information for sample embedding.

Subsequently, we adopt a simple linear interpolation function $\mathcal{G}_k(\cdot)$ to mix the node embeddings as formulated: 
\begin{equation}
\textbf{H}^{v_k} = \mathcal{G}_k(\textbf{H}) = \lambda\textbf{H} + (1-\lambda)\mathcal{S}_k(\textbf{H}),
\label{MIXUP_EMBEDDING}
\end{equation}
where $\textbf{H}^{v_k}$ denotes the $k$-th view of the node embedding and $\lambda=0.95$ is the interpolation rate. $\mathcal{S}_k(\cdot)$  is the shuffle function that randomly permutes the input of the function and output the same samples with a new order. 
As $\lambda=0.95$, the interpolation function could be regarded as an operation that introduces perturbation to the principal embedding $\textbf{H}$. 
Similar to Eq. (\ref{MIXUP_EMBEDDING}), the interpolated labels can be formulated as:
\begin{equation} 
\begin{aligned}
{\textbf{Y}^{v_k}} = \mathcal{G}_k(\textbf{Y}) = \lambda \textbf{Y} + (1 - \lambda) \mathcal{S}_k(\textbf{Y}).
\end{aligned}
\label{label_mix}
\end{equation}
In this manner, we construct two perturbations $\{\textbf{H}^{v_k}, {\textbf{Y}^{v_k}}\}$ as two different views of the principle sample batch in the latent space by mixing the node embeddings and the corresponding labels. Subsequently, we enhance the discriminative capability of the network by forcing the prediction model to change linearly between samples through the classification loss:

\begin{equation} 
\begin{aligned}
\mathcal{L}_C &= CE(\hat{\textbf{Y}}^{v_k}, {\textbf{Y}^{v_k}}),
\end{aligned}
\label{classification_loss}
\end{equation}
where $CE(\cdot)$ denotes the Cross-Entropy loss \cite{CE} and $\hat{\textbf{Y}}^{v_k}$ is the prediction of training data. According to \cite{mixuP, ICT}, in image classification applications, the decision boundaries are pushed far away from the class boundaries by enabling the network to recognize the interpolation operation. Through minimizing $\mathcal{L}_C$ in our paper, we can also acquire the larger-margin decision boundaries shown in Fig.\ref{t-sne}, thus alleviating the representation collapse problem.

\subsection{Correlation Reduction Module}
To further improve the discriminative capability of samples, we improve the cross-view consistency of the latent representation. Following this idea, as shown in the red box in Fig. \ref{OVERRALL_FIGURE}, we propose a correlation reduction module, which pulls together the same samples while pushing away different samples from two interpolation-perturbed views. In this way, our network is encouraged to learn more discriminative embeddings, thus avoiding the representation collapse problem.

Concretely, the process of correlation reduction is divided into three steps. First, we utilize the proposed graph interpolation module to construct two interpolation-perturbed views of node embeddings, i.e., $\textbf{H}^{v_1}$ and $\textbf{H}^{v_2}$ in Fig. \ref{OVERRALL_FIGURE}.

Second, the correlation matrix $\textbf{S} \in \mathbb{R}^{N \times N}$ across two interpolation-perturbed views is calculated as:

\begin{equation}
\textbf{S}_{ij}= \frac{\left(\textbf{H}^{v_1}_i\right) \left(\textbf{H}^{v_2}_j\right)^{\text{T}}}{{||\textbf{H}^{v_1}_i||}_2 {|| \textbf{H}^{v_2}_j ||_2}}, 
\label{correlation}
\end{equation}
where $\textbf{S}_{ij}$ is the cosine similarity between $i$-th node embedding of the first view ${{\textbf{H}^{v_1}}}$ and $j$-th node embedding of the second view ${{\textbf{H}^{v_2}}}$.

Furthermore, we force the correlation matrix $\textbf{S}$ to be equal to an identity matrix $\textbf{I}\in \mathbb{R}^{N\times N}$ by minimizing the information correlation reduction loss, which could be presented as:

\begin{equation}
\begin{aligned}
\mathcal{L}_R &= \frac{1}{N^2}\sum (\textbf{S}-\textbf{I})^2 \\
 &=  \frac{1}{N}\sum\limits_{i=1}^N \left(\textbf{S}_{ii}-1\right)^2
+
\frac{1}{N^2-N}\sum\limits_{i=1}^N \sum\limits_{j\ne i} \left(\textbf{S}_{ij}\right)^2.
\end{aligned}
\label{correlation_loss}
\end{equation}
In detail, the first term in Eq. (\ref{correlation_loss}) forces the diagonal elements of $\textbf{S}$ to 1, which indicates that the embeddings of each node are forced to agree with each other in two views. Besides, the second term in Eq. (\ref{correlation_loss}) makes the off-diagonal elements of $\textbf{S}$ to approach 0 so as to push away different nodes across two views.

\begin{algorithm}[!t]
\caption{Interpolation-based Correlation Reduction Network}
\label{ALGORITHM}
\textbf{Input}: An undirected graph $\mathcal{G}=({\mathbf{X}},\mathbf{A})$; Iteration number $t$; Hyper-parameters $\alpha$, $\lambda$. 
\\ \textbf{Output}: Class prediction $\hat{\textbf{Y}}$ and the trained network $\mathcal{F}(\cdot)$.

\begin{algorithmic}[1]

\FOR{$i=1$ to $t$}
\STATE{Encode the nodes with the feature extraction network $\mathcal{F}(\cdot)$ to obtain the node embeddings $\textbf{H}$;}
\STATE{Utilize the graph interpolation module to construct two interpolation-perturbed embeddings ${\textbf{H}}^{v_1}$ and ${\textbf{H}}^{v_2}$;}
\STATE Construct the interpolated labels ${\textbf{Y}}^{v_2}$ with Eq. \eqref{label_mix};
\STATE Calculate the classification loss \textit{$\mathcal{L}_{C}$} with Eq. \eqref{classification_loss};
\STATE Calculate the correlation matrix ${\textbf{S}}$ with Eq. \eqref{correlation};
\STATE Force ${\textbf{S}}$ to approximate an identity matrix and calculate information correlation reduction loss \textit{$\mathcal{L}_{R}$} with Eq. \ref{correlation_loss};
\STATE{Update the whole network by minimizing $\mathcal{L}$ in Eq. \eqref{LOSS}};
\ENDFOR
\STATE{Output the predicted classification result $\hat{\textbf{Y}}$.}
\STATE \textbf{return} $\hat{\textbf{Y}}$ and $\mathcal{F}(\cdot)$

\end{algorithmic}
\end{algorithm}

By this decorrelation operation, we enlarge the distance between different samples in the latent space while preserving the view-invariance latent feature of each sample, thus keeping cross-view consistent of latent representation. Consequently, our network is guided to learn more discriminative features about input samples and further avoid the collapsed representation.

\subsection{Loss Function}
The proposed method MGCN jointly optimizes two losses: the classification loss $\mathcal{L}_{C}$ and the information correlation reduction loss $\mathcal{L}_{R}$. In summary, the objective of MGCN is formulated as:
\begin{equation}
\mathcal{L} = \mathcal{L}_{C}+ \alpha \mathcal{L}_{R},
\label{LOSS}
\end{equation}
where $\alpha$ is a trade-off hyper-parameter. The detailed learning procedure of MGCN is illustrated in Algorithm \ref{ALGORITHM}.

\subsection{Theoretical Analysis}
In this subsection, we analyse the generalization bound of the proposed MGCN based on rademacher complexity \cite{rade}.

For the sake of convenience, we give the following notation. Let $G$ is the fixed graph with $n$ nodes. $S$ is the training set from an unknown distribution $\mathcal{D}$ where all the samples are i.i.d according to assumption of the previous researches \cite{pre}. 
 $\Gamma$ is the set of hyper-parameters.  $\mathcal{H}_\gamma$ is the distribution dependent hypothesis space corresponding the hyper-parameter $\gamma \in \Gamma$, which can be presented as $
\mathcal{H}_\gamma = \left \{ h_\gamma:\left ( \exists Z \in \mathcal{Z}  \right ) [h_\gamma = \mathcal{A}_\gamma(Z)]  \right \}$,
where $\mathcal{A}_\gamma$ is the algorithm to output the hypothesis $h_\gamma$ for a given training dataset $Z \in \mathcal{Z}$. For each $h_\gamma \in \mathcal{H}_\gamma$, $h_\gamma(\cdot,G)$ is the GNN network with $G$.  $\mathcal{R}^{\ell}_m{(\mathcal{H}_\gamma)}$ is the rademacher complexity of the set $\left \{(x,y)\mapsto \ell(h_\gamma(x,G), y):h_\gamma \in \mathcal{H}_\gamma\right \}$. $\mathbb{E}_{(x,y) \sim \mathcal{D}^n}[\ell(h_\gamma(x;G),y)] = \frac{1}{n}\sum_{i=1}^n \mathcal{L}(Z_i,h_\gamma)$. Let $c$ be the upper bound on the loss $\mathcal{L}$.

\begin{theorem}
For any $\delta>0$ and $\gamma \in \Gamma$, for all $h_\gamma \in \mathcal{H}_\gamma$, with the probability at least $1 - \delta/|\Gamma|$, we have:
\begin{equation}
\begin{aligned}
     &\mathbb{E}_{(x,y) \sim \mathcal{D}}[\ell(h_\gamma(x;G),y)]-\mathbb{E}_{(x,y) \sim \mathcal{D}^n}[\ell(h_\gamma(x;G),y)] \\
    &\le 2\sqrt{\frac{2\text{ln}\Pi_\mathcal{H_\gamma}(n)}{n}}+c(2+\alpha)\sqrt{\frac{\text{ln}(|\Gamma|/\delta)}{2n}}.
\end{aligned}
\end{equation}
\end{theorem}
where $\text{ln}\Pi_\mathcal{H}(n)$ denotes the growth function. For each fixed $\mathcal{H}_\gamma$, the generalization bound in Theorem 1 goes to zero since $\text{ln}\Pi_\mathcal{H_\gamma}(n)/{n} \rightarrow 0$ and $\text{ln}(|\Gamma|/\delta)/{n} \rightarrow 0$ when $n \rightarrow \infty$. In conclusion, the generation gap of our model is approximately $\mathcal{O}(1/{\sqrt{n}})$. Therefore, the generalization bound of the proposed MGCN is promised. 


Based on the Eq.\eqref{GPRGNN_ENCODING},\eqref{MIXUP_EMBEDDING},\eqref{label_mix},\eqref{classification_loss}, we formalize $\mathcal{L}_{C}$ as $\mathcal{L}_{C}(Z,h_\gamma(\cdot;G), \lambda)$. $\mathcal{L}(Z,h_\gamma)=\mathcal{L}_C(Z,h_\gamma)+\alpha \mathcal{L}_R(Z,h_\gamma)$. Denote $c$ as the upper bound on per-sample loss, i.e., $c \ge \ell(h_\gamma(x_i;G),y_i)$. Let $\gamma \in \Gamma$ be fixed.  Let $\mathbb{E}_{(x,y) \sim \mathcal{D}^n}[\ell(h_\gamma(x;G),y)] = \frac{1}{n}\sum_{i=1}^n \mathcal{L}(Z_i,h_\gamma)$. We define $\psi(Z) = sup_{h_\gamma \in \mathcal{H}_\gamma} \mathbb{E}_{(x,y) \sim \mathcal{D} }[\ell(h_\gamma(x;G),y)] - \mathbb{E}_{(x,y) \sim \mathcal{D}^n}[\ell(h_\gamma(x;G),y)]$. To prove the Theorem 1, we need the following two lemmas.

\begin{lemma}
\cite{rade} Let $\mathcal{F}$ be a class of real-valued function that map from $\mathcal{X}$ to  $[0,c]$. Let $\mathcal{D}$ be a probability distribution on $\mathcal{X} \times [0,c]$, and suppose that sample set $\mathbf{X} = \{x_1, x_2, \dots, x_n\}$ are chosen independently according to the distribution $\mathcal{D}$. For all $f \in \mathcal{F}$, with probability at least $1-\delta$, we have:

\begin{equation}
    \Phi(Z) \le 2 \mathcal{R}_n (\mathcal{F}) + c \sqrt{\frac{\text{ln}(1/\delta)}{2n}},
\end{equation}
where $\Phi(Z) = \text{sup}_{f \in \mathcal{F}} \mathbb{E}_{(x,y) \sim \mathcal{D} }[f] - \mathbb{E}_{(x,y) \sim \mathcal{D}^n}[f]$, $\mathcal{R}_n(\cdot)$ is the correspondent rademacher complexity.
\end{lemma}

\begin{lemma}
\cite{lemma_1} Let $\mathcal{H}$ be the hypothesis space. The Rademacher complexity $\mathcal{R}_n(\mathcal{H})$ and the growth function $\Pi_\mathcal{H}(n)$ have:
\begin{equation}
    \mathcal{R}_n(\mathcal{H}) \le \sqrt{\frac{2\text{ln}\Pi_\mathcal{H}(n)}{n}}.
\end{equation}
\end{lemma}

\begin{proof}
Firstly, we compute the upper bound on $\left | \psi(Z)-\psi(Z') \right | $. Here, $Z$ and $Z'$ denote the two training different datasets in one point of an arbitrary index $i_0$, which is formulated by:

\begin{equation}
\begin{aligned}
\psi(Z)-\psi(Z') &\le \text{sup}_{h_\gamma \in \mathcal{H}_\gamma} \mathcal{L}(Z, h_\gamma) -  \mathcal{L}(Z', h_\gamma)\\
&= \text{sup}_{h_\gamma \in \mathcal{H}_\gamma} (\mathcal{L}_C(Z, h_\gamma) - \mathcal{L}_C(Z', h_\gamma))\\
&+ \alpha(\mathcal{L}_R(Z, h_\gamma) - \mathcal{L}_R(Z', h_\gamma))
\end{aligned}
\label{bound_1}
\end{equation}
where $Z_i = Z'_i$ for all $i \neq i_0$ and $Z_{i0} \neq Z'_{i0}$. $\mathcal{L}_C$ and $\mathcal{L}_R$ denote the classification loss and information correlation reduction loss, respectively. For the first term:

\begin{equation}
\begin{aligned}
\mathcal{L}_{C}(Z, h_\gamma(\cdot,G),\lambda) - \mathcal{L}_{C}(Z', h_\gamma(\cdot,G),\lambda)\\
\le \frac{c(2n-1)}{n^2} \le \frac{2c}{n},
\end{aligned}
\end{equation}
where $Z$ and $Z'$ have $n^2$ terms and $2n-1$ different terms, each of which is bounded by the constant $c$. Therefore, for a fixed $G$ and a fixed $h_\gamma$, 

\begin{equation}
\begin{aligned}
\mathcal{L}_C(Z, h_\gamma) - \mathcal{L}_C(Z', h_\gamma) \le \frac{2c}{n}.
\end{aligned}
\label{L_c_bound}
\end{equation}

For the second term,

\begin{equation}
\begin{aligned}
\mathcal{L}_R(Z, h_\gamma) - \mathcal{L}_R(Z', h_\gamma) \le \frac{c}{n}. 
\end{aligned}
\label{L_r_bound}
\end{equation}

Since $\mathcal{L}_C$ is Cross-Entropy loss, the upper bound is 1. Besides, $S$ is calculated by the cosine similarity, the upper bound of $\mathcal{L}_R$ is 4. Moreover, we adopt the trade-off hyper-parameter $\alpha$ as 0.5. Therefore, the value of the upper bound $c$ is 3. Based on above proof, $\psi(Z)-\psi(Z') \le \frac{c(2+\alpha)}{n}$. We could obtain the similar bound $\psi(Z')-\psi(Z) \le \frac{c(2+\alpha)}{n}$. Therefore, for any $ \delta > 0$, with Lemma 2, at least the probability $1-\delta/|\Gamma|$:

\begin{equation}
\begin{aligned}
\psi(Z) \le 2\mathcal{R}_n(\mathcal{H}_\gamma) + c(2 + \alpha)\sqrt{\frac{ln(|\Gamma|/\delta)}{2n}}.  
\end{aligned}
\label{bound_1}
\end{equation}

Furthermore, with Lemma 3, we have:
\begin{equation}
\begin{aligned}
\psi(Z) \le 2\sqrt{\frac{2\text{ln}\Pi_\mathcal{H}(n)}{n}} + c(2 + \alpha)\sqrt{\frac{ln(|\Gamma|/\delta)}{2n}}.  
\end{aligned}
\label{bound_2}
\end{equation}

Therefore, we obtain that for any $\delta > 0$ and all $h_\gamma \in \mathcal{H}_\gamma$, with probability at least $1-\delta$:
\begin{equation}
\begin{aligned}
&\mathbb{E}_{(x,y) \sim \mathcal{D}}[\ell(h_\gamma(x,G),y)]- \mathbb{E}_{(x,y) \sim \mathcal{D}^n}[\ell(h_\gamma(x,G),y)]\\
&\le 2\sqrt{\frac{2\text{ln}\Pi_\mathcal{H}(n)}{n}} + c(2 + \alpha)\sqrt{\frac{\text{ln}(|\Gamma|/\delta)}{2n}}.  
\end{aligned}
\end{equation}

\end{proof}

\begin{table}[]
\centering
\caption{Dataset summary.}
\begin{tabular}{ccccc}
\hline
\textbf{Dataset}   & \textbf{Sample} & \textbf{Dimension} & \textbf{Edges} & \textbf{Classes} \\ \hline
\textbf{DBLP}      & 4507            & 334                & 7056           & 4                \\
\textbf{ACM}       & 3025            & 1870               & 26256          & 3                \\
\textbf{Photo}     & 7650            & 745                & 287326         & 8                \\
\textbf{Computers} & 13752           & 767                & 491722         & 10               \\
\textbf{Cora}      & 2708            & 1433               & 5429           & 7                \\
\textbf{Citeseer}  & 3327            & 3703               & 472            & 6                \\ \hline
\end{tabular}
\label{dataset_summary}
\end{table}

\section{Experiment}

\subsection{Datasets \& Metric}

To verify the effectiveness of our proposed method, extensive experiments have been conducted on six benchmark datasets, including DBLP \footnote{{\footnotesize{}{}\texttt{\footnotesize{}https://dblp.uni-trier.de}{\footnotesize{}}}}, ACM \footnote{{\footnotesize{}{}\texttt{\footnotesize{}https://dl.acm.org/}{\footnotesize{}}}}, AMAP \cite{GCC-LDA}, AMAC \cite{GCC-LDA}, CITESEER \footnote{{\footnotesize{}{}\texttt{\footnotesize{}http://citeseerx.ist.psu.edu/index}{\footnotesize{}}}}, and CORA \cite{AMAP_AMAC_CORA_CITESEER,DCRN}. Detailed dataset statistics are summarized in Table \ref{dataset_summary}. For fairness, we follow GPRGNN \cite{GPRGNN} and adopt the sparse splitting (2.5\% / 2.5\% / 95\% for train / validation / test) in the origin literature for all datasets. The classification performance is evaluated by the wide-used accuracy metric. 

\subsection{Experiment Setup}
All experiments are implemented with one NVIDIA 1080Ti GPU on PyTorch platform. To alleviate the influence of randomness, we run each method for 10 times and report the mean values with standard deviations. Besides, to all methods, we train them for 1000 epochs until convergence. For ACM and DBLP datasets, we adopt the code of compared methods and reproduce the results. For the performance of baselines on other datasets, we reported the corresponding values from GPRGNN \cite{GPRGNN} directly. In our proposed method, we adopt GPRGNN as our feature extraction backbone network, and our network is trained with the Adam optimizer \cite{adam}. Besides, the learning rate is set to 1e-3 for CITESEER, 5e-2 for DBLP, 2e-2 for CORA and AMAC, 1e-2 for ACM and AMAP, respectively. The interpolation rate $\lambda$ and the trade-off hyper-parameter $\alpha$ are set to 0.95 and 0.5, respectively. Moreover, the dimension $D$ is set to 64.

\begin{table*}[]
\centering
\caption{The average semi-supervised classification performance with mean±std on six datasets. The {\color[HTML]{FF0000} red} and {\color[HTML]{0000FF} blue} values indicate the best and the runner-up results, respectively.}
\scalebox{0.9}{
\begin{tabular}{cc|cccccccccccc}
\hline
\textbf{Method}        &               &  & \textbf{DBLP}                     &                         & \textbf{ACM}                      &                         & \textbf{Photo}                    &                         & \textbf{Computers}                &                         & \textbf{Citeseer}                 &                         & \textbf{Cora}                     \\ \hline
\textbf{MLP}           & \cite{MLP}             &  & 63.07±2.33                        &                         & 71.02±2.21                        &                         & 78.69±0.30                        &                         & 70.48±0.28                        &                         & 52.88±0.51                        &                         & 50.34±0.48                        \\
\textbf{GCN-Cheby}     & \cite{GCN_CHEBY}             &  & 60.48±0.00                        &                         & 79.98±3.07                        &                         & 90.09±0.28                        &                         & 82.41±0.28                        &                         & 65.67±0.38                        &                         & 71.39±0.51                        \\
\textbf{GCN}           & \cite{GCN}             &  & 67.64±0.38                        &                         & 84.95±0.21                        &                         & 90.54±0.21                        &                         & 82.52±0.32                        &                         & 67.30±0.35                        &                         & 75.21±0.38                        \\
\textbf{GraphSage}     & \cite{Graphsage}             &  & 29.49±0.03                        &                         & 37.65±0.01                        &                         & 90.51±0.25                        &                         & {\color[HTML]{0000FF} 83.11±0.23} &                         & 61.52±0.44                        &                         & 70.89±0.54                        \\
\textbf{APPNP}         & \cite{PPNP}             &  & 67.75±0.44                        &                         & 74.61±0.67                        &                         & 91.11±0.26                        &                         & 81.99±0.26                        &                         & {\color[HTML]{0000FF} 68.59±0.30} &                         & 79.41±0.38                        \\
\textbf{JK-Net}        & \cite{JKNet}             &  & 64.51±0.53                        &                         & 81.20±0.11                        &                         & 87.70±0.70                        &                         & 77.80±0.97                        &                         & 60.85±0.76                        &                         & 73.22±0.64                        \\
\textbf{GAT}           & \cite{GAT}             &  & 68.58±0.42                        &                         & 83.88±0.35                        &                         & 90.09±0.27                        &                         & 81.95±0.38                        &                         & 67.20±0.46                        &                         & 76.70±0.42                        \\
\textbf{SGC}           & \cite{SGC}             &  & 53.66±2.15                        &                         & 72.99±2.96                        &                         & 83.80±0.46                        &                         & 76.27±0.36                        &                         & 58.89±0.47                        &                         & 70.81±0.67                        \\
\textbf{GPRGNN}        & \cite{GPRGNN}             &  & 67.84±0.30                        &                         & 80.93±2.26                        &                         & {\color[HTML]{0000FF} 91.93±0.26} &                         & 82.90±0.37                        &                         & 67.63±0.38                        &                         & {\color[HTML]{0000FF} 79.51±0.36} \\
\textbf{AdaGCN}        & \cite{adagcn}            &  & {\color[HTML]{0000FF} 69.70±1.35} &                         & {\color[HTML]{333333} 86.09±1.99} &                         & 46.44±3.25                        &                         & 39.71±0.77                        &                         & 62.58±1.44                        &                         & 62.41±1.84                        \\ \hline
\textbf{DGI}           & \cite{DGI}            &  & {\color[HTML]{343434} 68.90±1.34} &                         & {\color[HTML]{000000} 81.26±1.48} &                         & 83.10±0.50                        &                         & 75.90±0.60                        &                         & 65.43±2.94                        &                         & 73.74±1.43                        \\
\textbf{GCA}           & \cite{GCA}            &  & {\color[HTML]{343434} 60.11±1.94} &                         & 79.23±1.73                        &                         & 89.98±1.28                        &                         & 81.86±1.80                        &                         & 66.25±3.94                        &                         & 74.49±3.70                        \\
\textbf{GRACE}         & \cite{GRACE}            &  & 68.88±0.04                        &                         & 85.93±0.56                        &                         & 90.60±0.03                        &                         & 72.76±0.02                        &                         & 66.54±0.01                        &                         & 78.62±0.62                        \\
\textbf{MVGRL}         & \cite{MVGRL}            &  & 67.89±0.34                        &                         & 83.78±0.27                        &                         & 79.37±0.03                        &                         & 70.22±0.02                        &                         & 67.98±0.05                        &                         & 78.06±0.07                        \\
\textbf{BGRL}          & \cite{bgrl}            &  & 68.48±0.56                        &                         & 86.04±0.67                        &                         & 90.78±0.89                        &                         & 81.40±0.25                        &                         & 67.78±0.85                        &                         & 78.87±0.41                        \\ \hline
\textbf{MixupForGraph} & \cite{GRAPH_MIXUP_1}            &  & {\color[HTML]{343434} 68.51±0.78} &                         & 86.24±0.62                        &                         & 89.87±0.10                        &                         & 77.30±2.10                        &                         & 57.41±0.33                        &                         & 67.11±0.63                        \\
\textbf{GraphMix}      & \cite{graphmix}            &  & {\color[HTML]{343434} 67.98±0.36} &                         & 85.75±0.55                        &                         & 90.14±1.81                        &                         & 79.24±0.57                        &                         & 66.24±0.74                        &                         & 78.54±0.32                        \\
\textbf{GraphMixup}    & \cite{graphmixup}            &  & 68.48±1.52                        &                         & {\color[HTML]{0000FF} 86.44±1.25} &                         & 90.74±0.78                        &                         & 81.68±0.42                        &                         & 67.53±0.87                        &                         & 77.16±1.12                        \\ \hline
\textbf{MGCN}          & \textbf{Ours} &  & {\color[HTML]{FF0000} 70.60±0.76} & {\color[HTML]{FF0000} } & {\color[HTML]{FF0000} 87.88±0.54} & {\color[HTML]{FF0000} } & {\color[HTML]{FF0000} 92.64±0.24} & {\color[HTML]{FF0000} } & {\color[HTML]{FF0000} 83.99±0.90} & {\color[HTML]{FF0000} } & {\color[HTML]{FF0000} 69.18±0.43} & {\color[HTML]{FF0000} } & {\color[HTML]{FF0000} 80.89±0.95} \\ \hline
\end{tabular}}
\label{COMPARE_RESULT}
\end{table*}

\subsection{Performance Comparison}

{To demonstrate the superiority of our method, we conduct performance comparison experiments for our proposed MGCN and 18 baselines, including classical semi-supervised graph methods, unsupervised contrastive methods, and Mixup-enhanced methods. 
\begin{itemize}
    \item Classical semi-supervised graph methods (\cite{MLP,GCN_CHEBY, GCN, Graphsage, JKNet, GAT, SGC, GPRGNN, PPNP}) propagate the supervision information from the labeled data to the unlabeled data according to the linkages within the adjacent matrix as guidance for network training.
    \item We report the results of the unsupervised contrastive methods (\cite{DGI, GCA, GRACE, MVGRL,bgrl}), which design auxiliary tasks for discriminative information exploitation.
    \item Mixup-based methods (\cite{GRAPH_MIXUP_1,graphmix,graphmixup}) improve the robustness and discriminative capability of the aggregated sample embedding over the labeled samples. 
\end{itemize}
}

From these results in Table \ref{COMPARE_RESULT}, we observe and analyze as follows. 1) It could be observed that the classical GCN-based methods are not comparable with our proposed MGCN. Taking the results on CORA dataset, MGCN exceeds GCN \cite{GCN} by 5.68\%. This is because these methods would suffer from the representation collapse problem caused by the inaccurate connections within data in the adjacency matrix. 2) Our MGCN consistently outperforms other contrastive learning methods. We conjecture that those methods fail to exploit abundant intuitive information within the unlabeled data, thus achieving sub-optimal performance. 3) Compared with the Mixup-enhanced methods, MGCN achieves better classification performance. The reason is that those graph-mix-based methods do not consider the contrastive learning method to improve the discriminative capacity in the semi-supervised node classification task. {Overall, MGCN alleviate collapsed representations by improving the discriminative capability of the latent space from two aspects. Firstly, we proposed a graph interpolation to force the prediction model to change linearly between samples. The the margin of decision boundaries can be enlarged, thus improving the discriminative of the network. Besides, the proposed correlation reduction mechanism further improves the discriminative capability of the embeddings by keeping the cross-view consistency of the latent representations. Consequently, the proposed MGCN alleviates collapsed representations and achieves the top-level performance on six datasets.}

{Our proposed MGCN aims to alleviate collapsed representations in graph node classification field. The advantages of MGCN can be summarized as follows: 1) Overcoming irregularity and connectivity challenges: The irregularity and connectivity of graph data make it difficult to directly utilize an interpolation-based strategy. To overcome this challenge, we introduce an interpolation-based framework that enables data augmentation on embeddings and associated labels. 2) Simplified augmentation strategy: Instead of relying on complex graph augmentation techniques such as graph diffusion or attribute shuffle, we employ an easily implemented interpolation-based strategy to manipulate the embeddings. This approach proves effective in enlarging the decision boundaries and improving the model's discriminative capacity.}

\begin{figure*}
\centering
\scriptsize
\begin{minipage}{0.19\linewidth}
\centerline{\includegraphics[width=\textwidth]{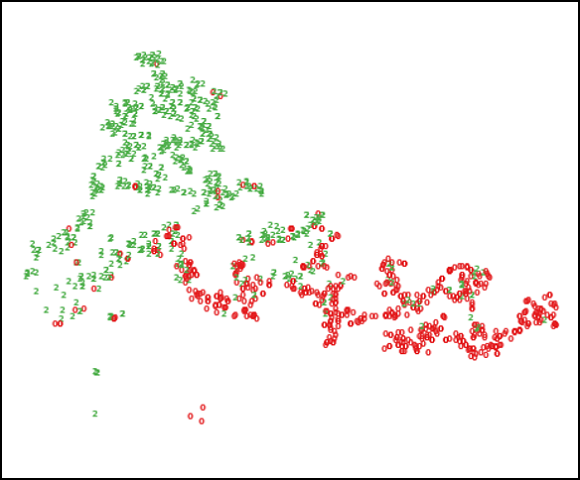}}
\vspace{3pt}
\centerline{\includegraphics[width=\textwidth]{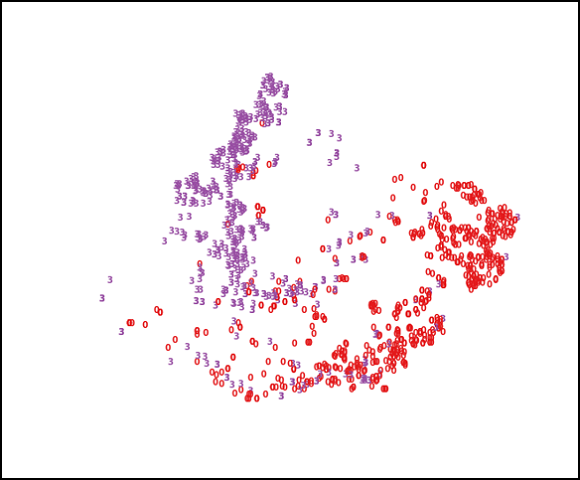}}
\vspace{3pt}
\centerline{ChebNet}
\end{minipage}
\begin{minipage}{0.19\linewidth}
\centerline{\includegraphics[width=\textwidth]{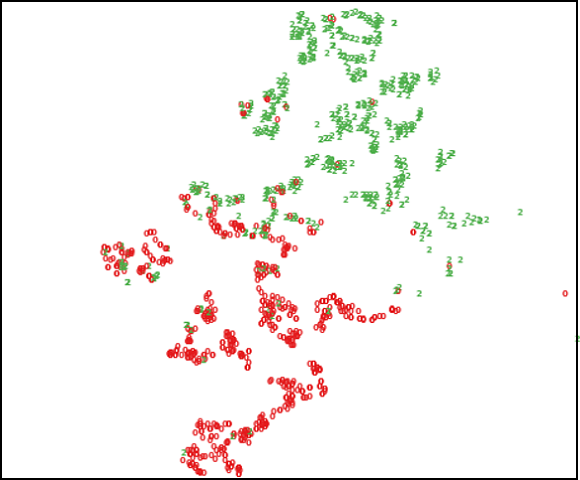}}
\vspace{3pt}
\centerline{\includegraphics[width=\textwidth]{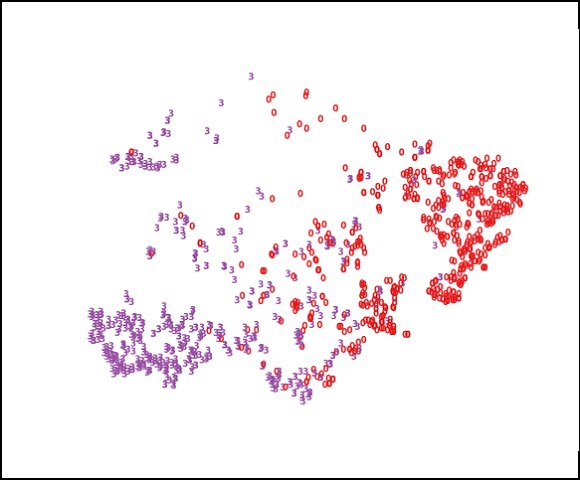}}
\vspace{3pt}
\centerline{GCN}
\end{minipage}
\begin{minipage}{0.19\linewidth}
\centerline{\includegraphics[width=\textwidth]{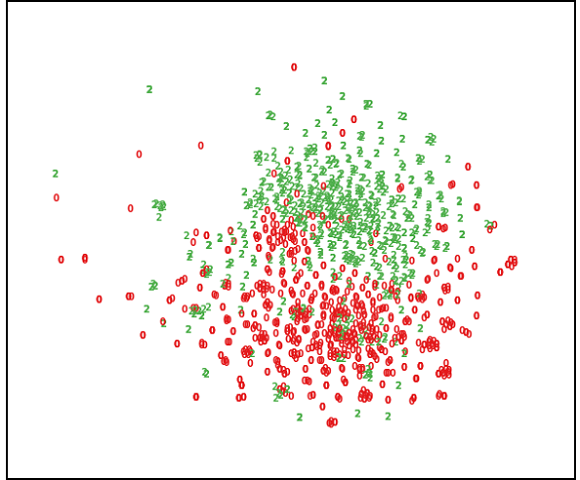}}
\vspace{3pt}
\centerline{\includegraphics[width=\textwidth]{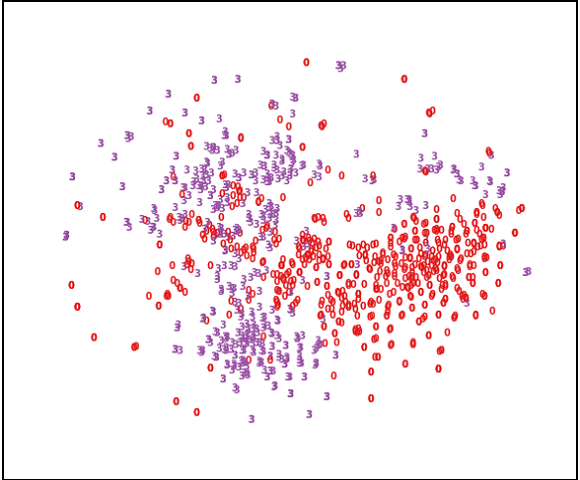}}
\vspace{3pt}
\centerline{MVGRL}
\end{minipage}
\begin{minipage}{0.19\linewidth}
\centerline{\includegraphics[width=\textwidth]{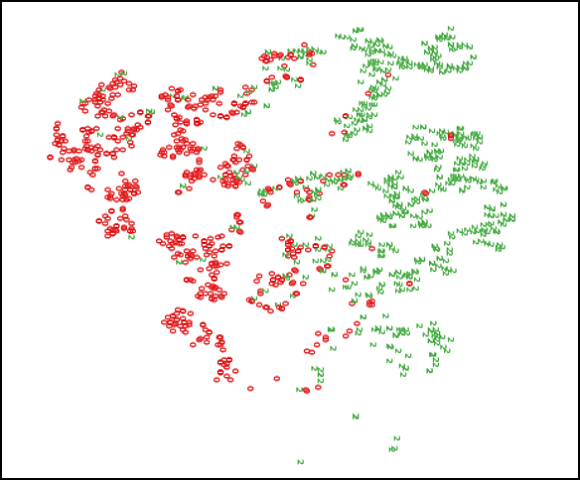}}
\vspace{3pt}
\centerline{\includegraphics[width=\textwidth]{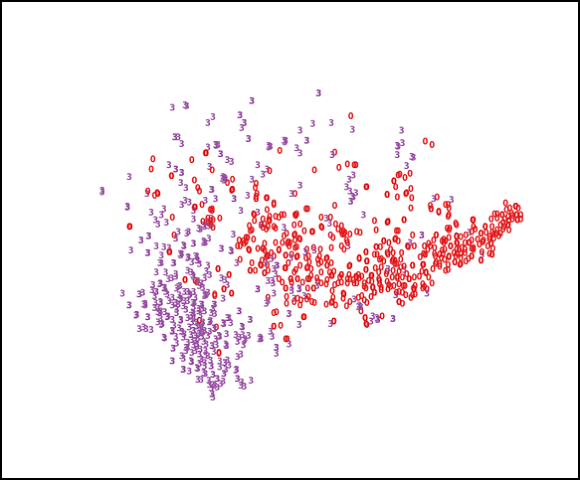}}
\vspace{3pt}
\centerline{GPRGNN}
\end{minipage}
\begin{minipage}{0.19\linewidth}
\centerline{\includegraphics[width=\textwidth]{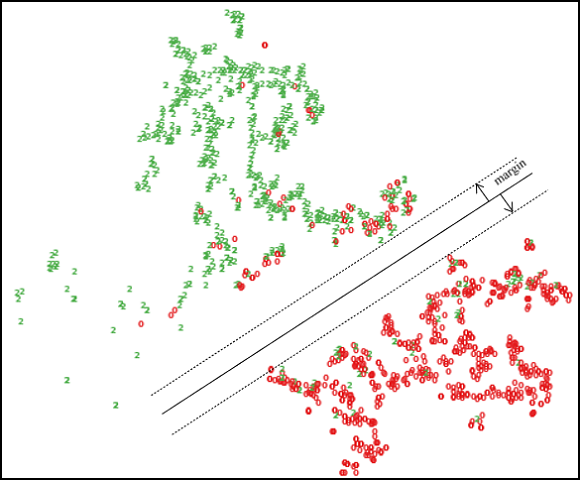}}
\vspace{3pt}
\centerline{\includegraphics[width=\textwidth]{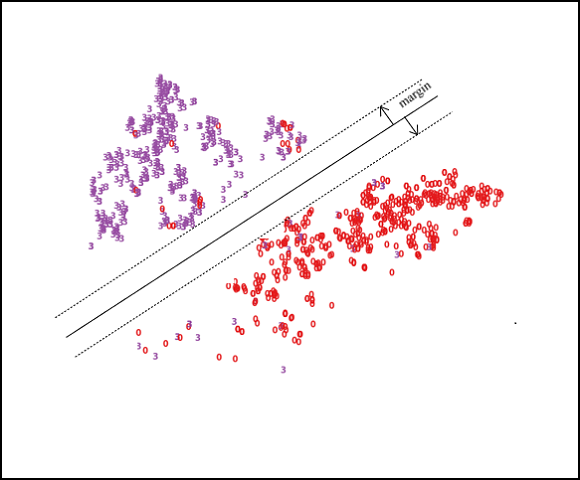}}
\vspace{3pt}
\centerline{Ours}
\end{minipage}
\caption{$t$-SNE visualization of seven methods on two datasets. The first row and second row correspond to ACM and DBLP, respectively.}
\label{t-sne}  
\end{figure*}

\begin{figure*}
\begin{minipage}{0.33\linewidth}
\centerline{\includegraphics[width=1\textwidth]{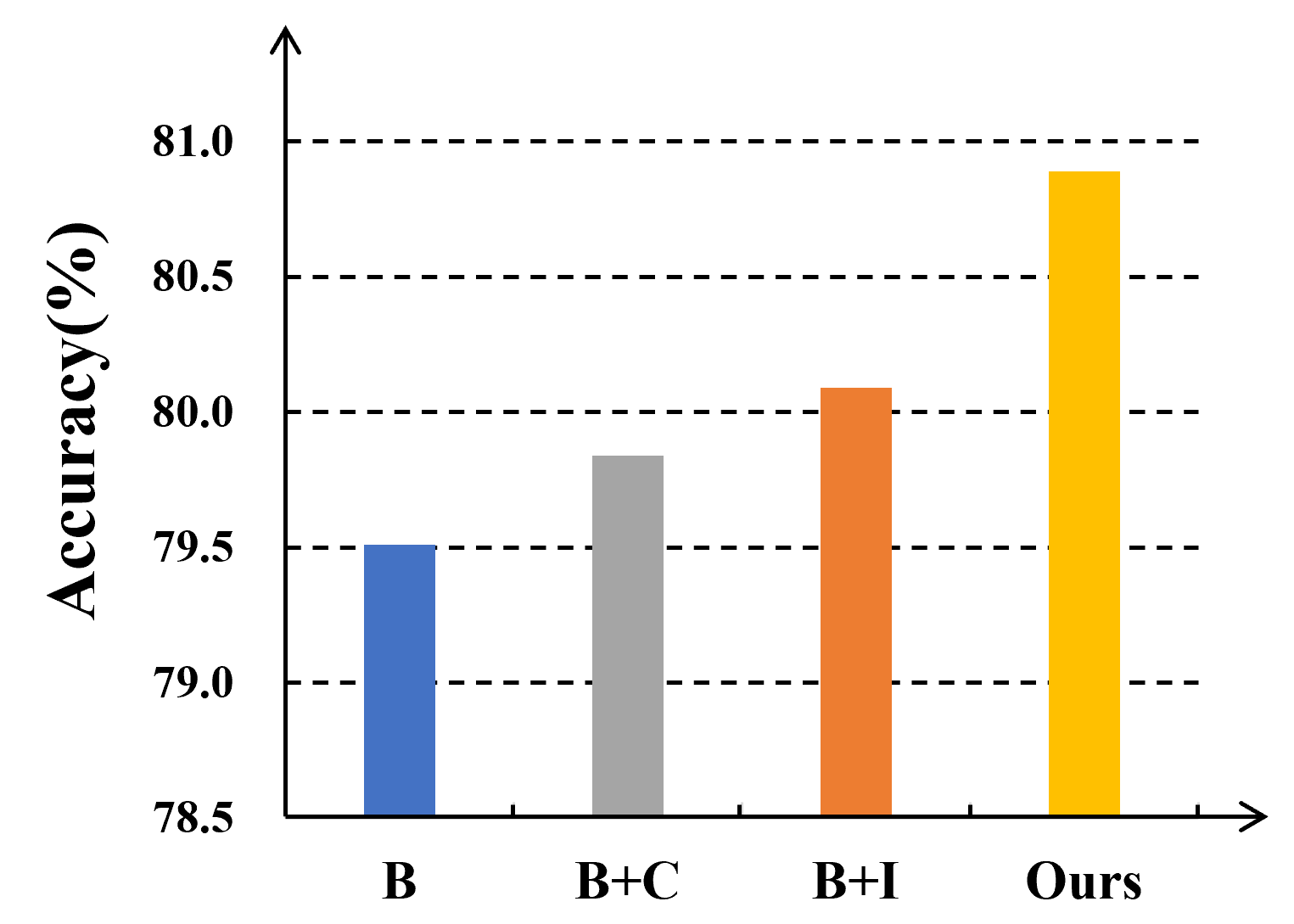}}
\vspace{3pt}
\centerline{\textbf{CORA}}
\vspace{3pt}
\centerline{\includegraphics[width=1\textwidth]{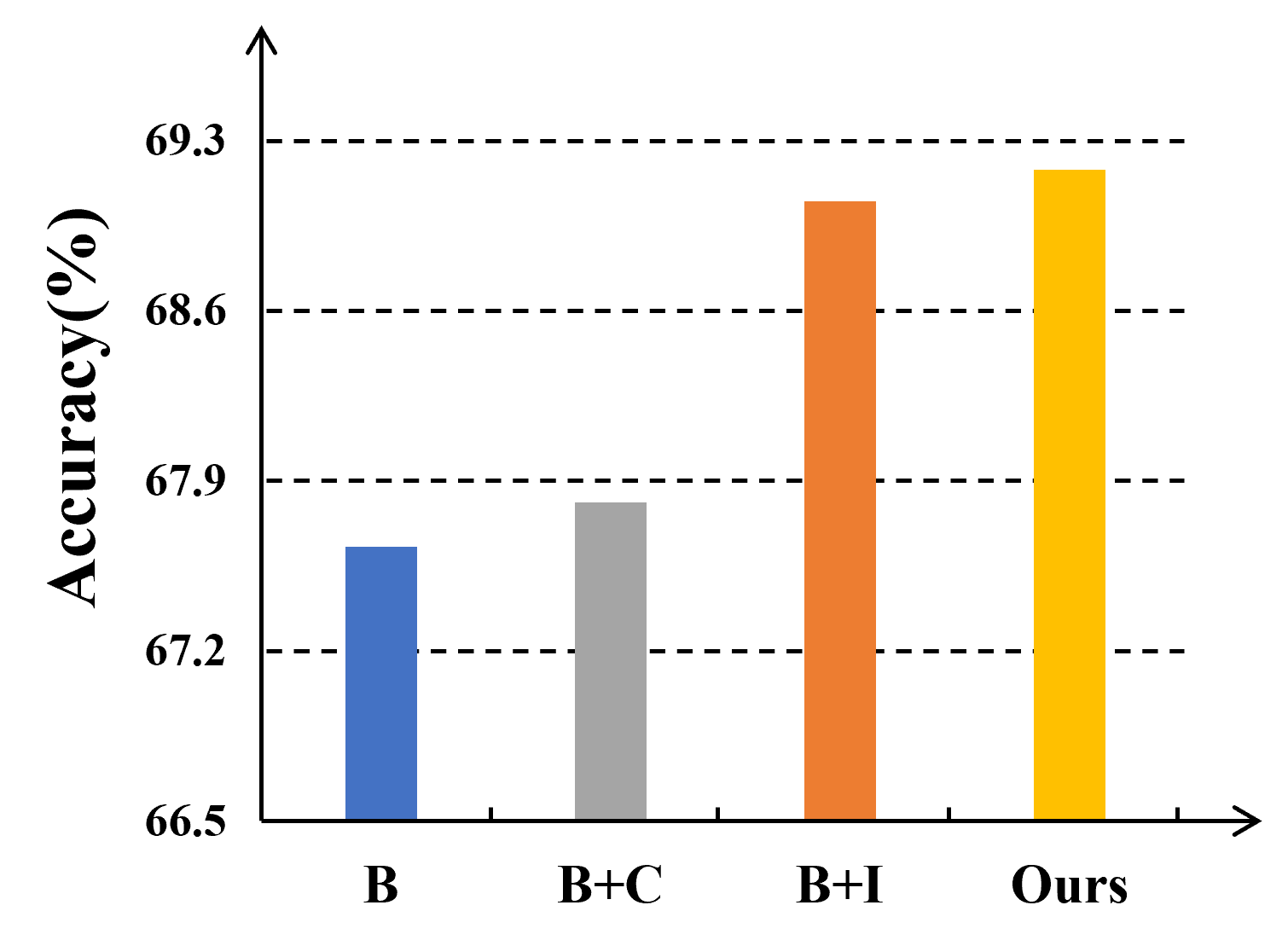}}
\vspace{3pt}
\centerline{\textbf{CITESEER}}
\end{minipage}
\begin{minipage}{0.33\linewidth}
\centerline{\includegraphics[width=1\textwidth]{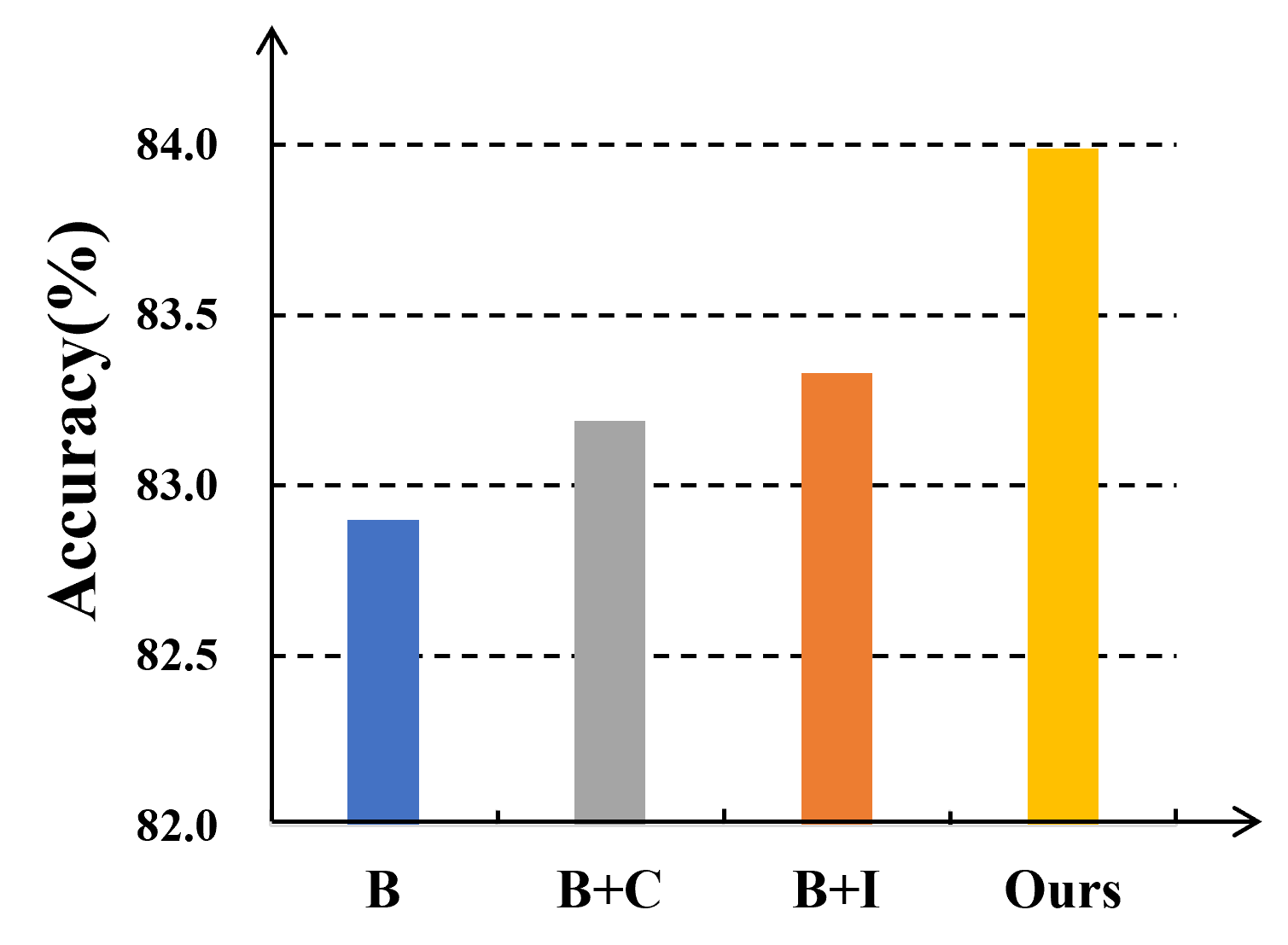}}
\vspace{3pt}
\centerline{\textbf{AMAC}}
\vspace{3pt}
\centerline{\includegraphics[width=1\textwidth]{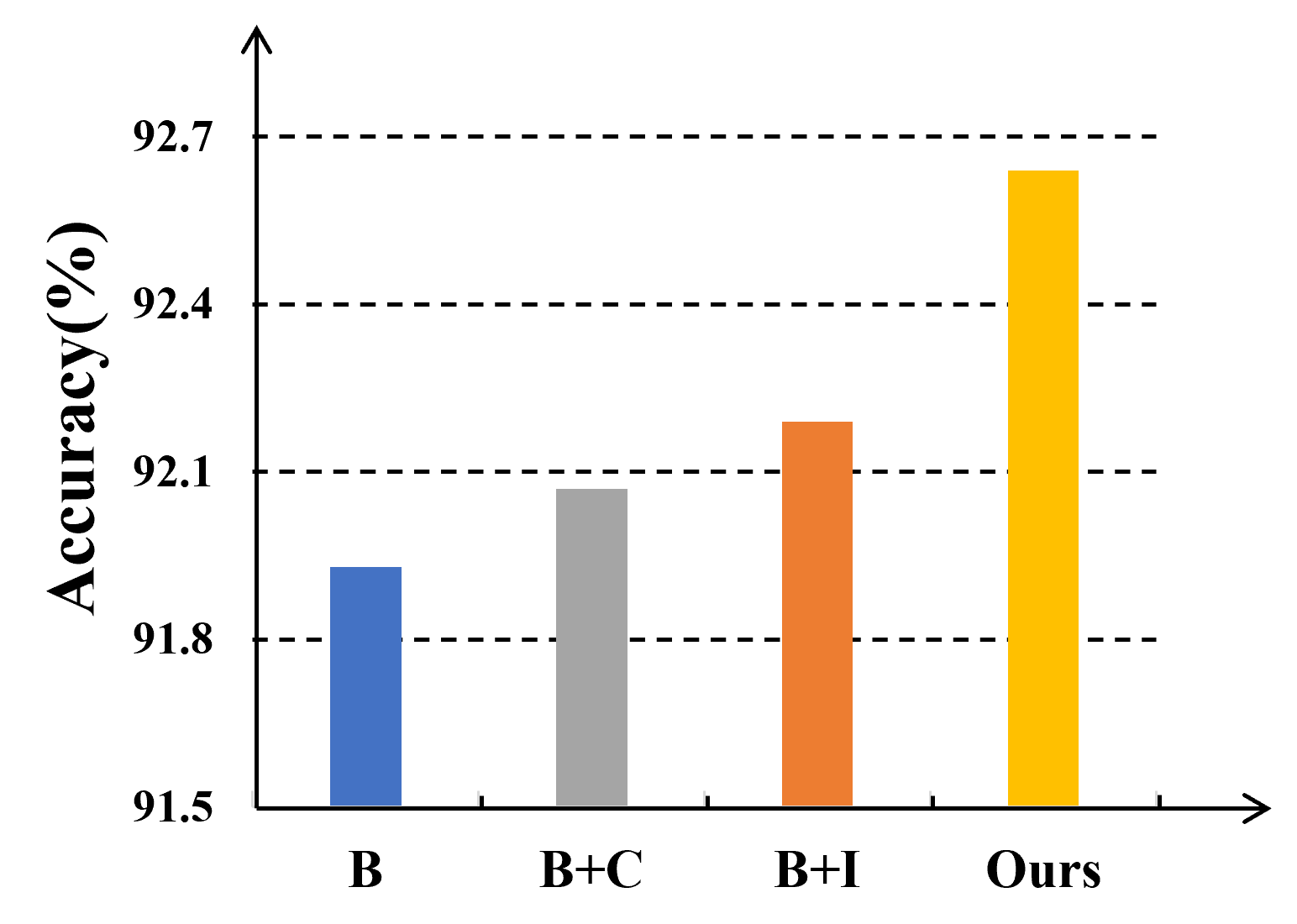}}
\vspace{3pt}
\centerline{\textbf{AMAP}}
\end{minipage}
\begin{minipage}{0.33\linewidth}
\centerline{\includegraphics[width=1\textwidth]{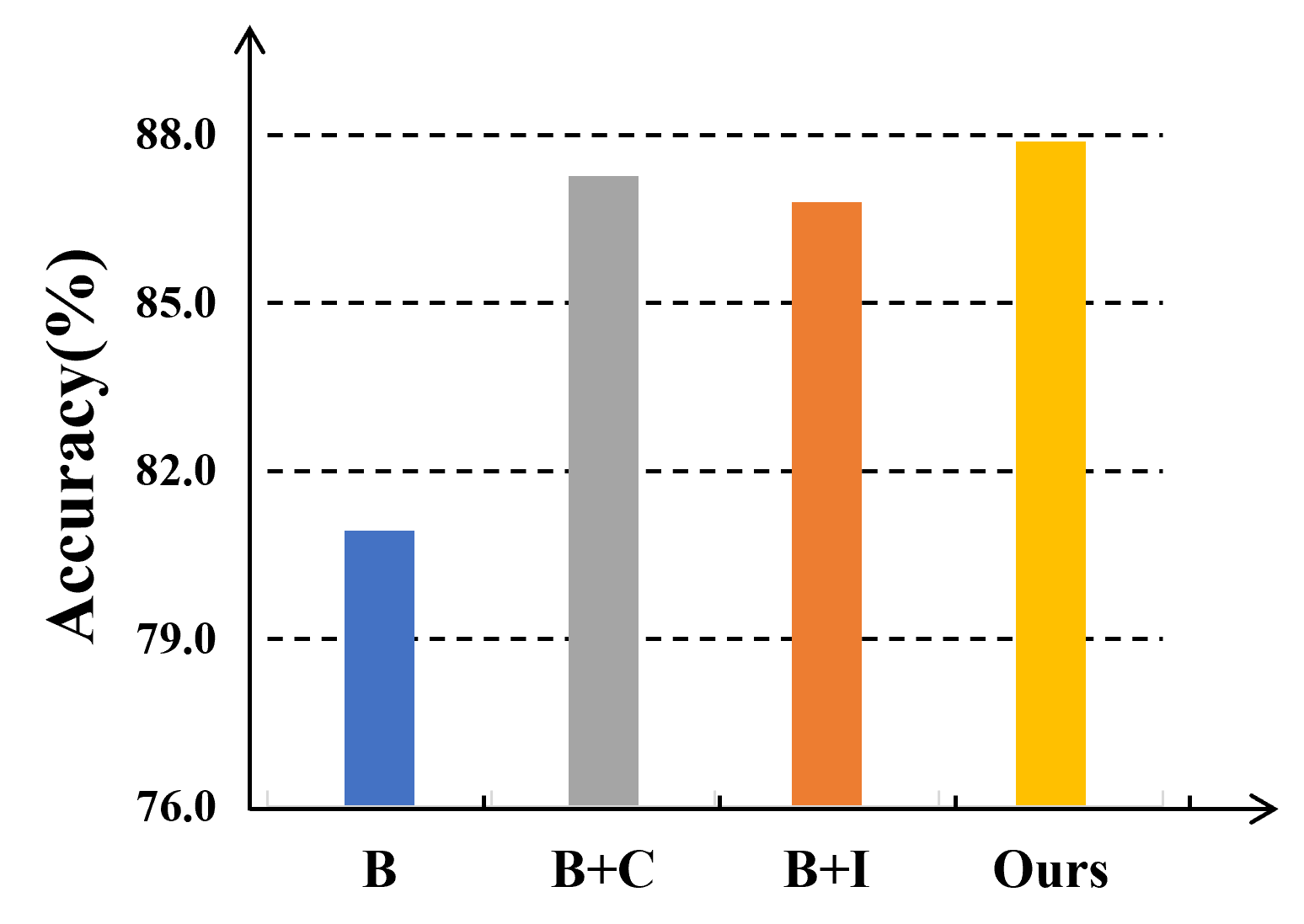}}
\vspace{3pt}
\centerline{\textbf{ACM}}
\vspace{3pt}
\centerline{\includegraphics[width=1\textwidth]{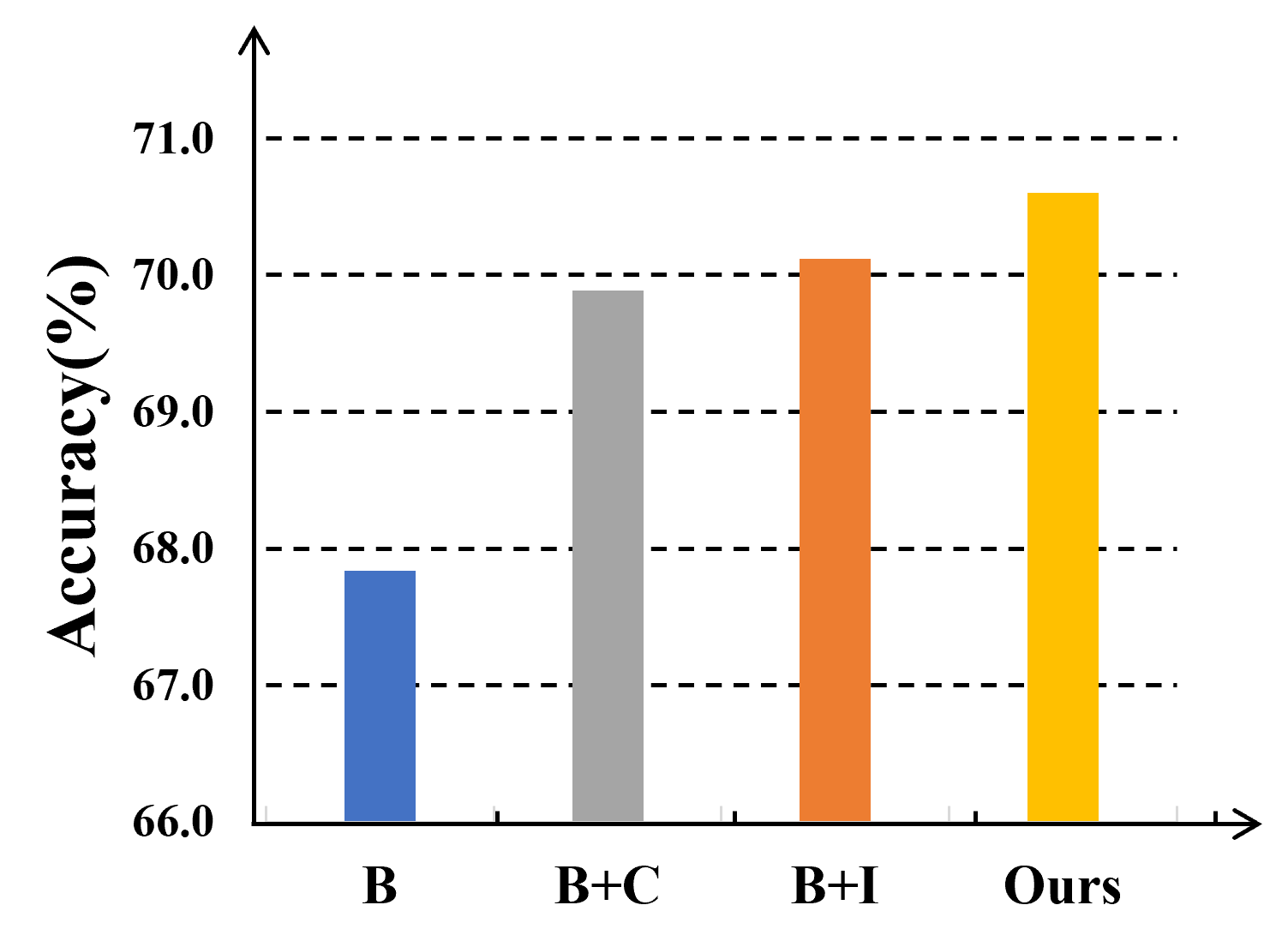}}
\vspace{3pt}
\centerline{\textbf{DBLP}}
\end{minipage}
\caption{{Ablation comparisons of the proposed modules on six datasets.  ``B'', ``B+I'', ``B+C'' and ``Ours'' denote the baseline, the baseline with graph interpolation module, correlation reduction module and both, respectively.}}
\label{ABLATION_MODULE}
\end{figure*}
\vspace{5pt}

\begin{figure*}
\centering
\includegraphics[scale=0.5]{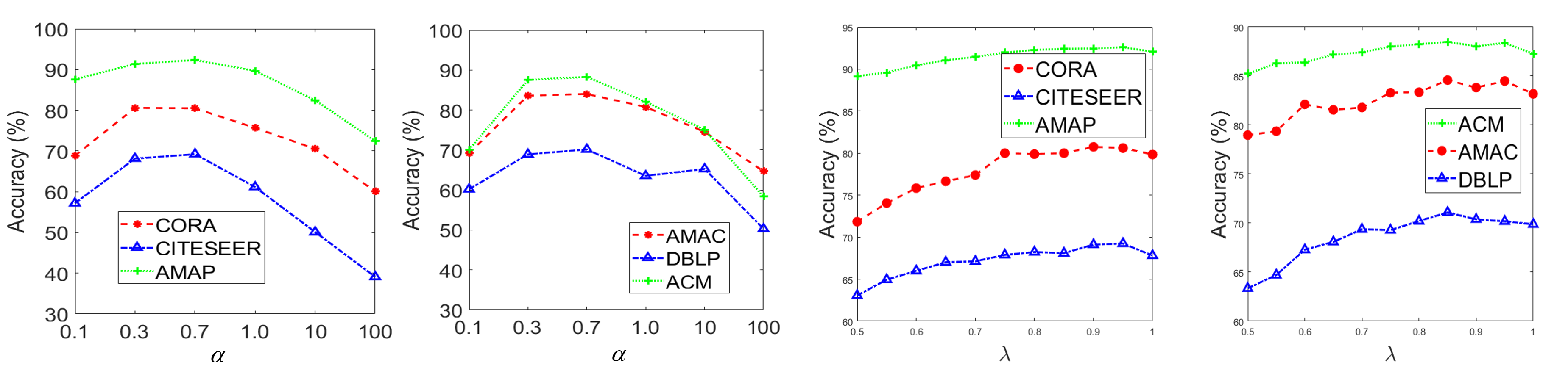} 
\vspace{5pt}
\caption{{Testing of the effectiveness and sensitivity of hyper-parameter $\alpha$ and $\lambda$. The result perturbation with the variation of the two parameters on all six datasets are illustrated in the figures.}}
\label{ablation_study_a}
\end{figure*}


\subsection{Transferring Modules to Other Methods}
To further investigate the effectiveness and the generality of our proposed modules, we transfer the graph interpolation module and correlation reduction module to five baselines including GCN-Cheby \cite{GCN_CHEBY}, GCN \cite{GCN}, APPNP\cite{PPNP}, JK-Net \cite{JKNet}, GAT \cite{GAT}. Table \ref{Transfer} reports the performance of the five methods with their variants on DBLP, ACM, CITESEER, and CORA dataset. Here, we denotes the baseline and the baseline with the two proposed modules as B and B-O, respectively.

From these results, we observed that, enhanced by our proposed modules, the baselines significantly achieve better performance. Specifically, our modules improve the classification accuracy of GCN by 4.79\% on DBLP, 0.82\% on ACM, 1.23\% on CITESEER, 2.49\% on CORA, respectively. The reason is that the two proposed modules enhance the discriminative capability of samples by conducting the interpolation-based augmentations in the latent space and improving the cross-view consistency of the node representations. In this manner, the baselines alleviate the collapsed representation, thus achieving better classification performance.

\begin{table}[]
\centering
\caption{{Time cost comparisons of the training process. All results are measured in seconds with ten runs.}}
\scalebox{0.9}{
\begin{tabular}{c|cccccc}
\hline
\textbf{Method}        & \textbf{Cora} & \textbf{Citeseer} & \textbf{AMAP} & \textbf{AMAC} & \textbf{ACM} & \textbf{DBLP} \\ \hline
\textbf{GCN}           & 8.56s         & 9.33s             & 21.26s        & 37.77s        & 8.45s        & 8.2s          \\
\textbf{APPNP}         & 11.45s        & 14.25s            & 12.75s        & 38.75s        & 11.39s       & 11.34s        \\
\textbf{JKNet}         & 14.9s         & 14.71s            & 15.01s        & 57.56s        & 14.79s       & 14.71s        \\
\textbf{ChebyNet}      & 13.9s         & 22.49s            & 46.37s        & 90.65s        & 20.89s       & 10.42s        \\
\textbf{MixupForGraph} & 13.35s        & 20.66s            & 120.1s        & 59.68s        & 17.23s       & 11.47s        \\
\textbf{GPRGNN}        & 3.06s         & 3.06s             & 3.5s          & 6.05s         & 10s          & 10.37s        \\ \hline
\textbf{MGCN}          & 7.18s         & 9.55s             & 15.5s         & 49.17s        & 7.06s        & 6.6s          \\ \hline
\end{tabular}}
\label{time_compare}
\end{table}

\subsection{Time Cost and Memory Cost}

In this subsection, we implement time and memory cost experiments to demonstrate the effectiveness of the proposed MGCN. Specifically, we test the training time of MGCN with five baselines on six datasets. For fairness, we train all algorithms with 1000 epochs. The results are shown in Table \ref{time_compare}. From the results, we observe that the training time of MGCN is comparable with other state-of-the-art methods. The reasons we analyze are as follows: 1) Following GPRGNN \cite{GPRGNN}, instead of using GCN we adopt generalized pagerank (GPR) to jointly extract node feature and topological information. This operation effectively reduces time consumption. 2) MGCN utilize Eq.\ref{correlation} as loss function. Compared with commonly contrastive loss function, e.g., temperature-scaled cross-entropy
loss (NT-Xent) \cite{SIMCLR,simgrace,xihong}, the loss function in MGCN is more efficient. The detailed description is shown in \ref{loss_effect}.

\begin{figure*}[!t]
\centering 
\includegraphics[width = 0.75\linewidth]{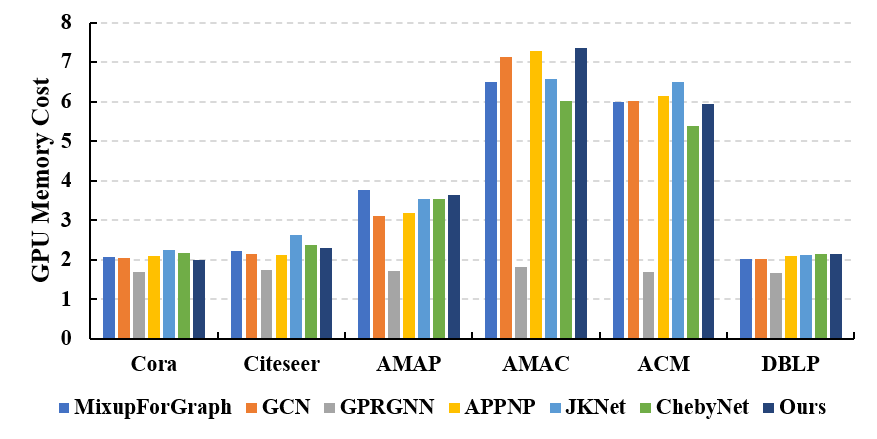}
\caption{GPU memory costs of seven methods on six datasets.}
\label{gpu_cost}  
\end{figure*}

Moreover, we conduct experiments to test GPU memory costs of our proposed MGCN with six methods (i.e., MixupForGraph \cite{GRAPH_MIXUP_1}, GCN \cite{GCN}, GPRGNN \cite{GPRGNN}, APPNP \cite{PPNP}, JKNet \cite{JKNet}, ChebNet \cite{GCN_CHEBY}) on six datasets. From the results in Fig. \ref{gpu_cost}, we observe that the memory costs of our MGCN are comparable with other algorithms. We summarize the reasons as follows: 1) Following GPRGNN \cite{GPRGNN}, MGCN adopts the simple network architecture. 2) Compared with temperature-scaled cross-entropy
loss (NT-Xent) \cite{SIMCLR,xihong} in other contrastive learning algorithms, MGCN utilizes MSE loss to reduce memory costs.

\begin{table*}[]
\centering
\caption{Transferring our proposed modules to other models on  four datasets. 'B' and 'B-O' represent the baseline and the baseline with our method, respectively.
Boldface letters are used to mark the best results.}
\scalebox{0.75}{
\begin{tabular}{c|cc|cc|cc|cc|cc}
\hline
\multirow{2}{*}{\textbf{Dataset}} & \multicolumn{2}{c|}{\textbf{GCN-Cheby}} & \multicolumn{2}{c|}{\textbf{GCN}} & \multicolumn{2}{c|}{\textbf{APPNP}} & \multicolumn{2}{c|}{\textbf{JKNet}} & \multicolumn{2}{c}{\textbf{GAT}}          \\ \cline{2-11} 
                                  & \textbf{B}     & \textbf{B-O}           & \textbf{B}  & \textbf{B-O}        & \textbf{B}   & \textbf{B-O}         & \textbf{B}   & \textbf{B-O}         & \textbf{B}          & \textbf{B-O}        \\ \hline
\textbf{DBLP}                     & 60.48±0        & \textbf{63.52±1.46}    & 67.64±0.38  & \textbf{72.43±0.62} & 67.84±0.30   & \textbf{68.50±0.78}  & 64.51±0.53   & \textbf{66.97±0.49}  & 68.58±0.42          & \textbf{69.00±1.84} \\ \hline
\textbf{ACM}                      & 79.98±3.07     & \textbf{83.02±1.03}    & 84.95±0.21  & \textbf{85.77±1.33} & 74.61±0.67   & \textbf{83.71±1.78}  & 81.20±0.11   & \textbf{85.53±1.22}  & \textbf{83.88±0.35} & 83.18±2.93          \\ \hline
\textbf{CITESEER}                 & 65.67±0.38     & \textbf{66.52±0.65}    & 67.30±0.35  & \textbf{68.53±0.59} & 68.59±0.30   & \textbf{70.12±0.97}  & 60.85±0.76   & \textbf{64.88±1.00}  & 67.20±0.46          & \textbf{68.54±0.38} \\ \hline
\textbf{CORA}                     & 71.39±0.51     & \textbf{72.95±1.06}    & 75.21±0.38  & \textbf{77.70±0.44} & 79.41±0.38   & \textbf{79.53±0.37}  & 73.22±0.64   & \textbf{75.45±1.69}  & 76.70±0.42          & \textbf{77.25±3.25} \\ \hline
\end{tabular}}
\label{Transfer}
\end{table*}

\begin{figure*}
\centering
\begin{minipage}{0.235\linewidth}
\centerline{\includegraphics[width=\textwidth]{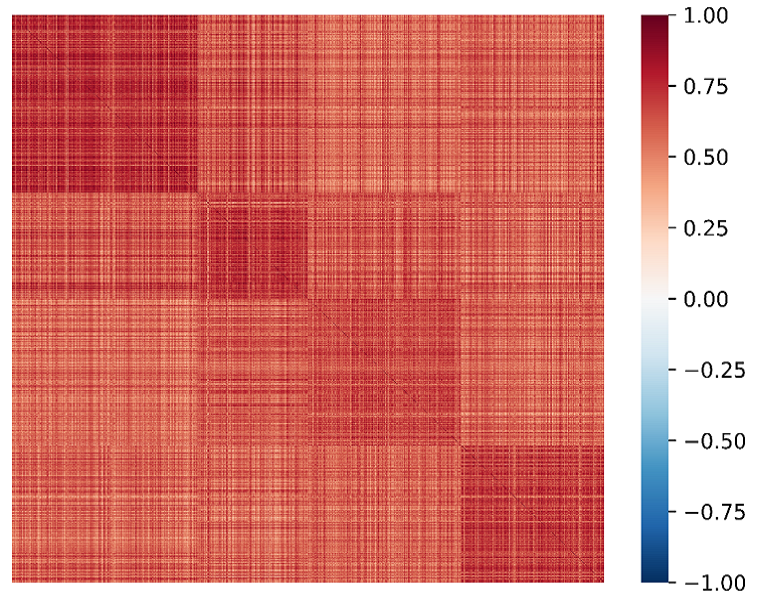}}
\vspace{3pt}
\centerline{\includegraphics[width=\textwidth]{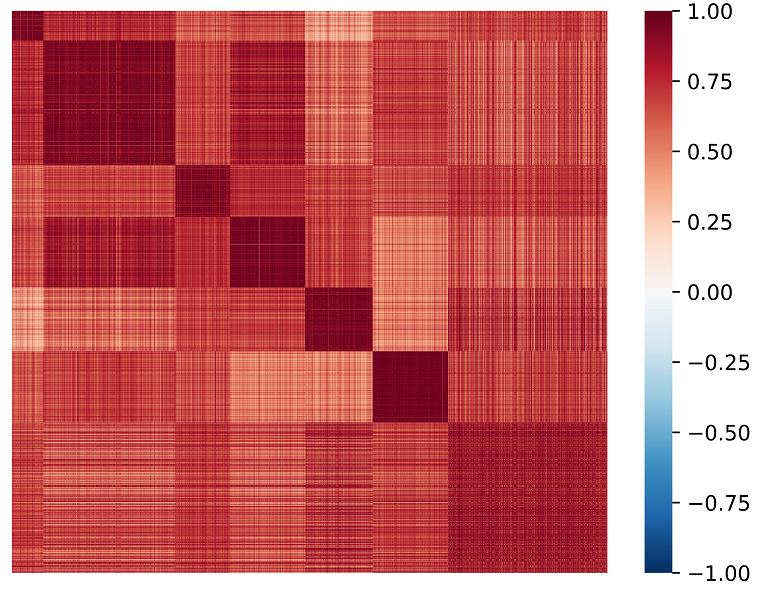}}
\vspace{3pt}
\centerline{GCN}
\end{minipage}
\begin{minipage}{0.235\linewidth}
\centerline{\includegraphics[width=\textwidth]{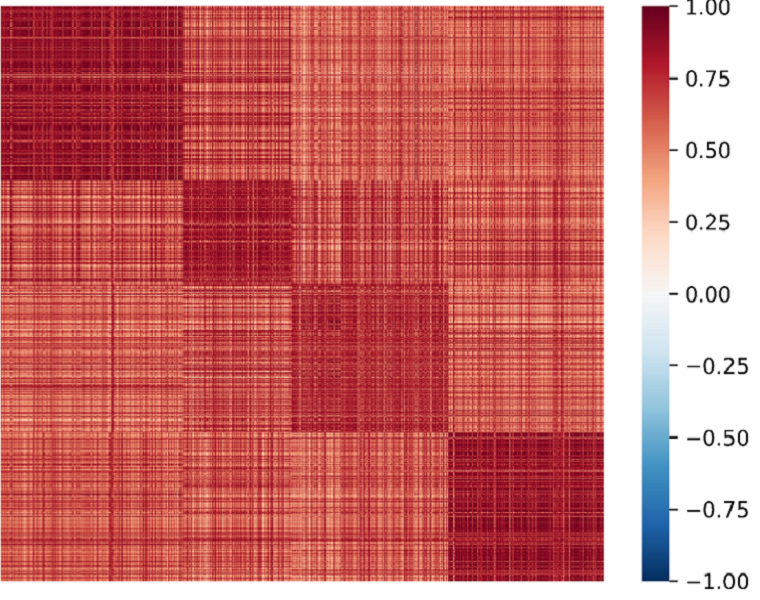}}
\vspace{3pt}
\centerline{\includegraphics[width=\textwidth]{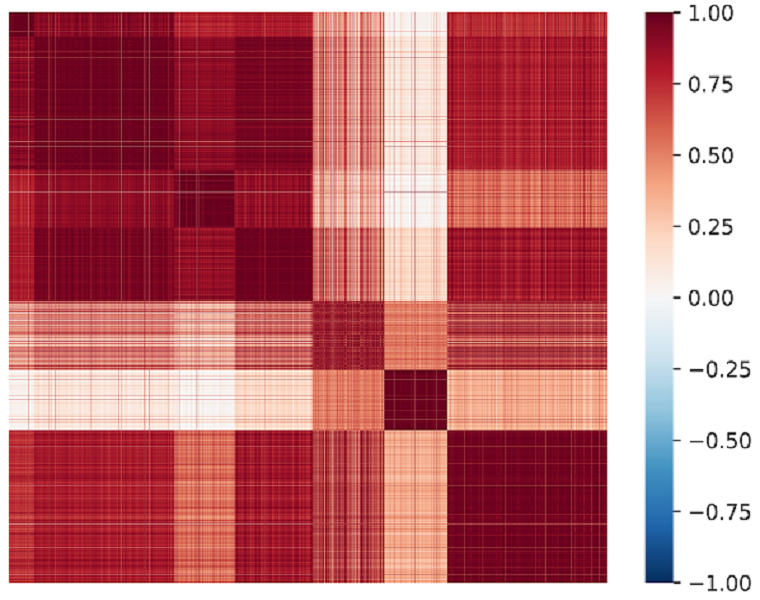}}
\vspace{3pt}
\centerline{JK-Net}
\end{minipage}
\begin{minipage}{0.235\linewidth}
\centerline{\includegraphics[width=\textwidth]{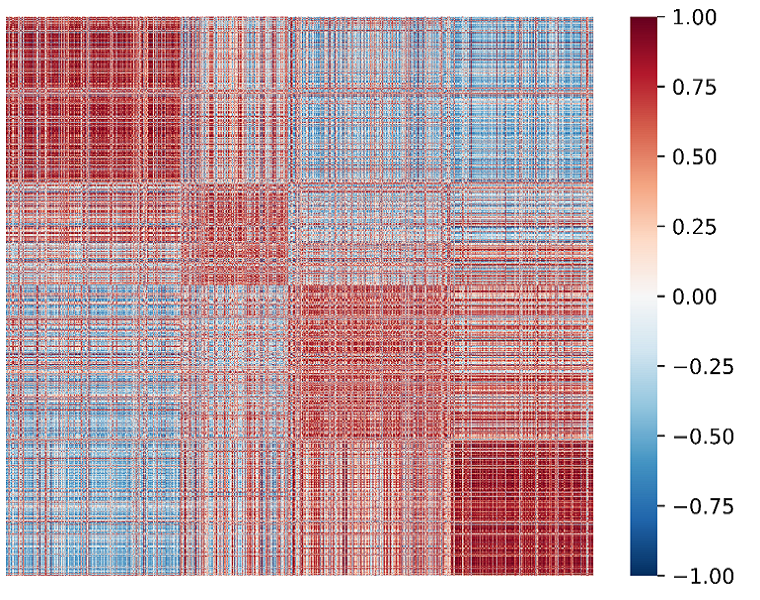}}
\vspace{3pt}
\centerline{\includegraphics[width=\textwidth]{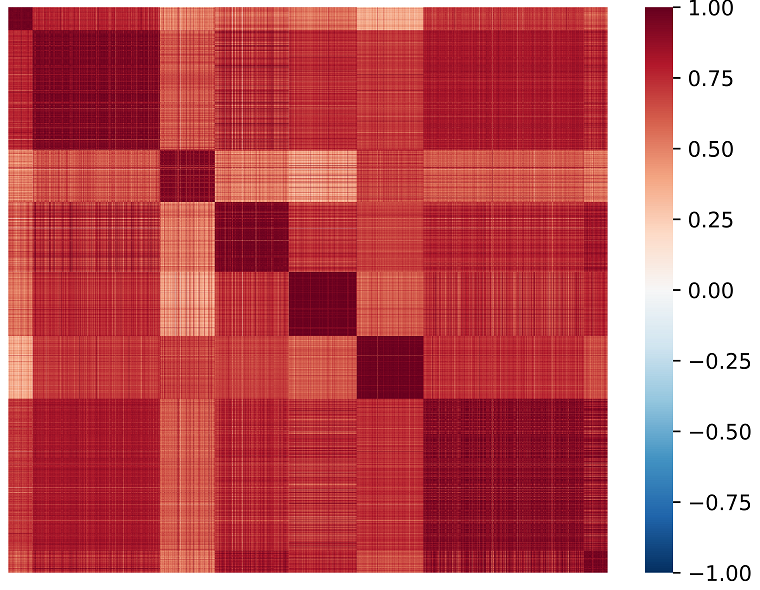}}
\vspace{3pt}
\centerline{GPRGNN}
\end{minipage}
\begin{minipage}{0.235\linewidth}
\centerline{\includegraphics[width=\textwidth]{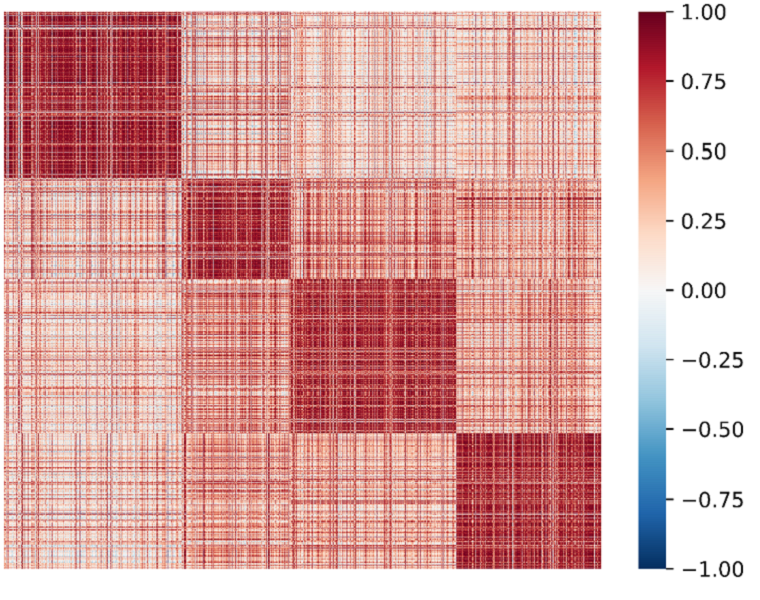}}
\vspace{3pt}
\centerline{\includegraphics[width=\textwidth]{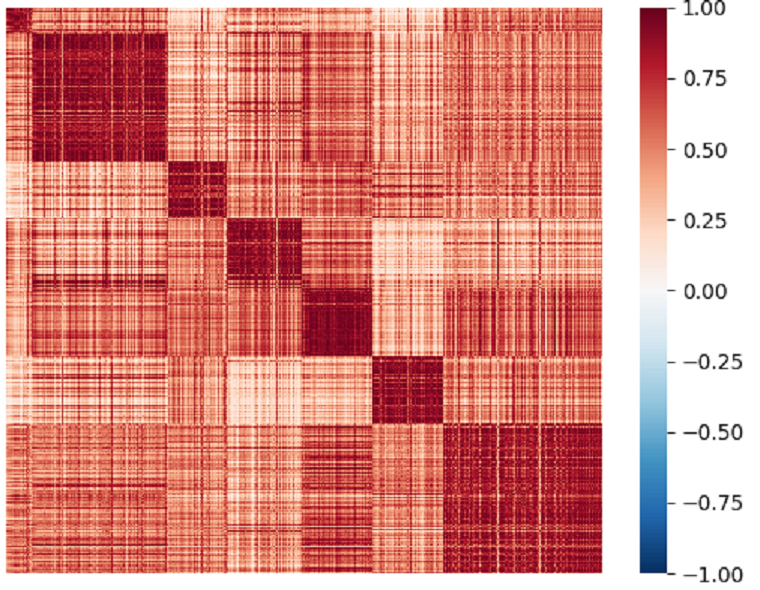}}
\vspace{2pt}
\centerline{Ours}
\end{minipage}
\caption{Visualization of sample similarity matrices of the network output on two datasets. The first row and second row correspond to DBLP and AMAP, respectively. The higher value (\textcolor{red}{red}) indicates that embeddings are more similar, thus easy leading to representation collapsing. The lower value (\textcolor{blue}{blue}) denotes that the embeddings are less similar.}
\label{similarity}  
\end{figure*}

\subsection{Ablation Studies}
In this section, we first conduct ablation studies to verify the effectiveness of the proposed modules, and then we analyze the robustness of MGCN to the hyper-parameters. Last, we conduct experiments to verify the effectiveness of our proposed loss function.

\subsubsection{Effectiveness of the Proposed Modules}
To investigate the effectiveness of the proposed graph interpolation module and correlation reduction module, extensive ablation studies are conducted in Fig. \ref{ABLATION_MODULE}. Here, we adopt GPNGNN\cite{GPRGNN} as ``Baseline''. Besides, ``B'', ``B+I'', ``B+C'' and ``Ours'' denote the baseline, the baseline with graph interpolation module, correlation reduction module and both, respectively. From these results, we have observed as follows. 1) Compared with ``Baseline'', ``B+I'' has about 1.81\% performance improvement on average of six datasets since the proposed graph interpolation module enlarges the margin of decision boundaries by forcing the prediction model to change linearly between samples. 2) Benefited from the correlation reduction module, the classification performance is improved. Taking the result on DBLP dataset for example, ``B+C'' exceeds ``Baseline'' by 2.05\%. This demonstrates that the correlation reduction module improves the discriminative capability of samples by keeping the cross-view consistency of the latent representations. 3) Moreover, better performance of ``Ours'' indicates that both proposed modules are effective to guide the network to learn more discriminative latent features.

\begin{figure*}
\centering
\scriptsize
\begin{minipage}{0.19\linewidth}
\centerline{\includegraphics[width=\textwidth]{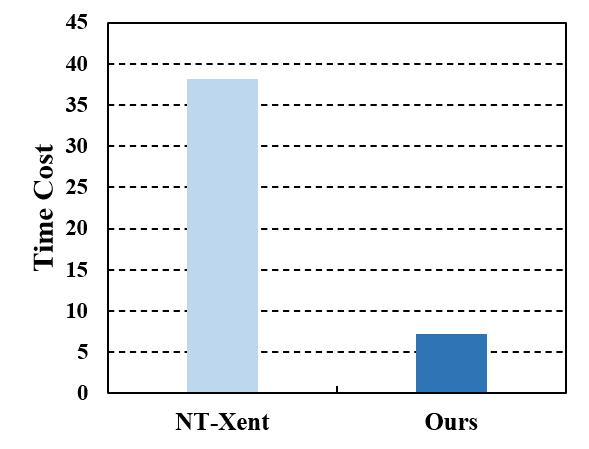}}
\vspace{3pt}
\centerline{\includegraphics[width=\textwidth]{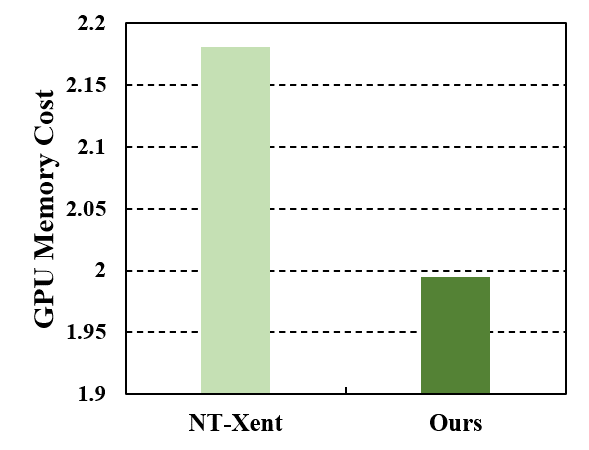}}
\vspace{3pt}
\centerline{Cora}
\end{minipage}
\begin{minipage}{0.19\linewidth}
\centerline{\includegraphics[width=\textwidth]{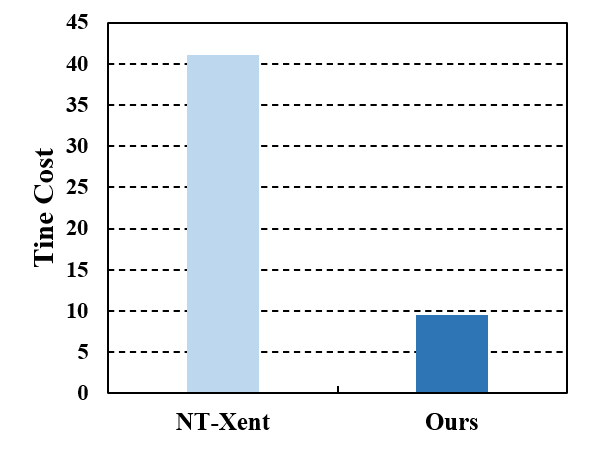}}
\vspace{3pt}
\centerline{\includegraphics[width=\textwidth]{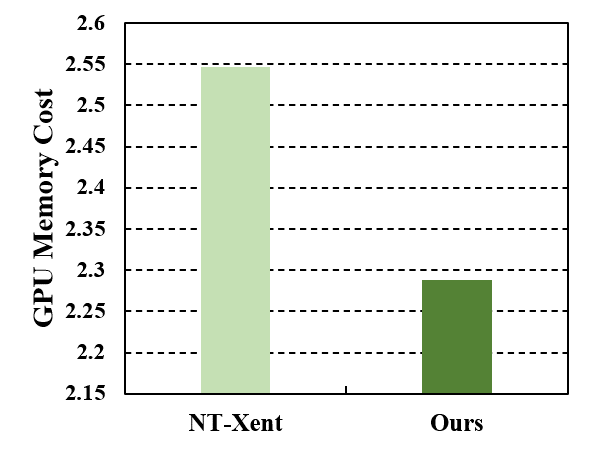}}
\vspace{3pt}
\centerline{Citeseer}
\end{minipage}
\begin{minipage}{0.19\linewidth}
\centerline{\includegraphics[width=\textwidth]{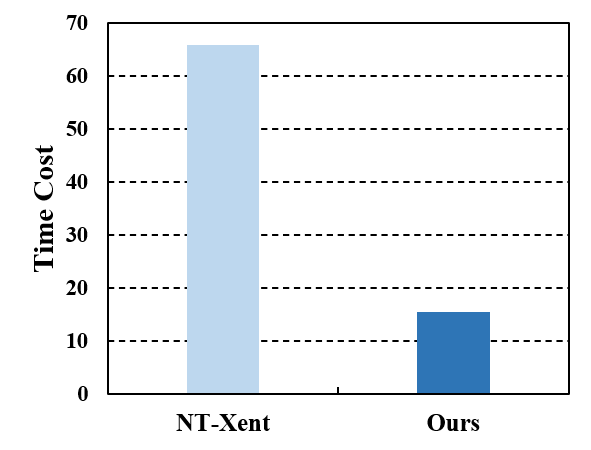}}
\vspace{3pt}
\centerline{\includegraphics[width=\textwidth]{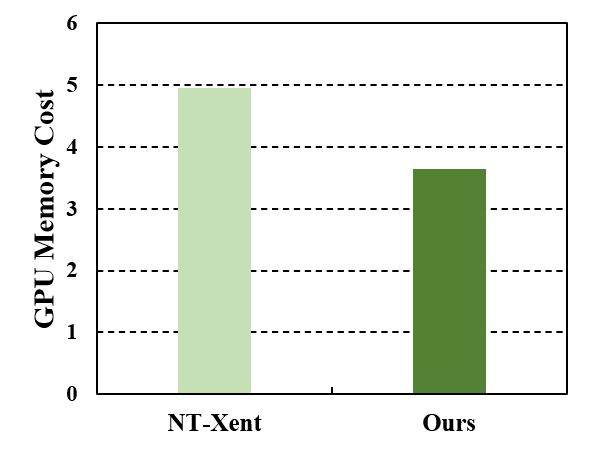}}
\vspace{3pt}
\centerline{AMAP}
\end{minipage}
\begin{minipage}{0.19\linewidth}
\centerline{\includegraphics[width=\textwidth]{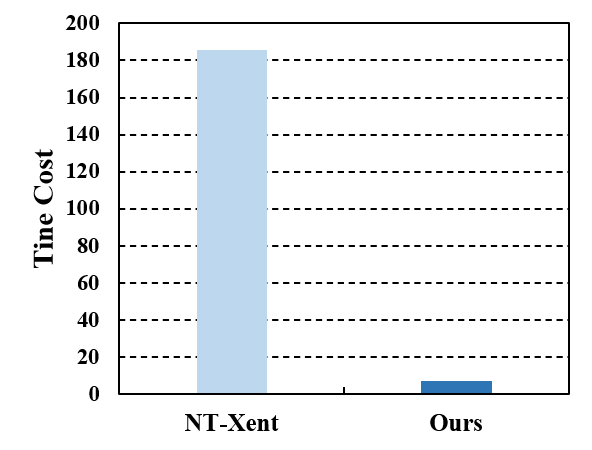}}
\vspace{3pt}
\centerline{\includegraphics[width=\textwidth]{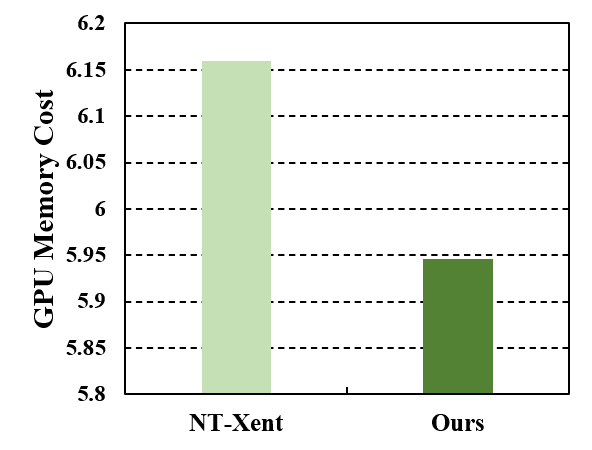}}
\vspace{3pt}
\centerline{ACM}
\end{minipage}
\begin{minipage}{0.19\linewidth}
\centerline{\includegraphics[width=\textwidth]{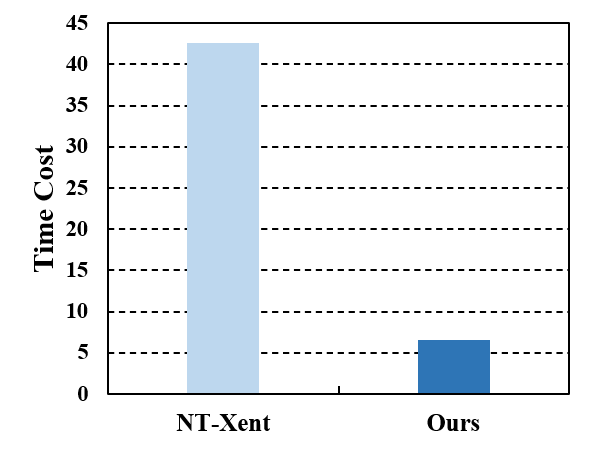}}
\vspace{3pt}
\centerline{\includegraphics[width=\textwidth]{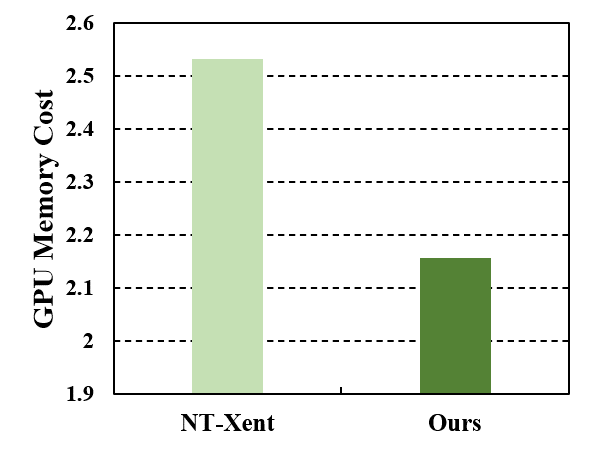}}
\vspace{3pt}
\centerline{DBLP}
\end{minipage}
\caption{Time costs and GPU memory costs of NT-Xent and our loss functions on five datasets. NT-Xent denotes temperature-scaled cross-entropy loss.}
\label{time_gpu}  
\end{figure*}

\subsubsection{Hyper-parameter Analysis}
Furthermore, we investigate the robustness of our proposed method to the hyper-parameters on six datasets. Specifically, to the trade-off hyper-parameter $\alpha$, we conduct ablation studies as shown in Fig. \ref{ablation_study_a} (a). From these results, we observe that the classification accuracy will not fluctuate greatly when $\alpha \in [0.3, 0.7]$. This demonstrates that our model MGCN is insensitive when $\alpha$ varying in a tiny range. Besides, the accuracy of semi-supervised node classification with different values of the interpolation rate $\lambda$ are illustrated in Fig. \ref{ablation_study_a} (b). It's observed that the performance of MGCN is decreased when $\lambda$ is about less than 0.95 since $\lambda$ controls the perturbation to the principal embedding $\textbf{H}$. It is worth mentioning that $\lambda$ is set as 0.95 in all experiments.

\subsubsection{Effectiveness of the Loss Function} \label{loss_effect}

To demonstrate the effectiveness of our proposed loss function, we have conducted experiments to test the GPU memory costs and time costs. Specifically, we compare with NT-Xent loss \cite{SIMCLR} on five datasets, including CORA, CITESEER, AMAP, ACM, and DBLP. From Fig. \ref{time_gpu}, we could observe that the time cost of our loss function is less than NT-Xent. Moreover, the GPU memory costs are also comparable.

\subsection{Visualization Experiment}
\subsubsection{$t$-SNE Visualization of Classification Results}
To intuitively show the superiority of MGCN, we visualize the distribution of the node embeddings $\textbf{H}$ learned by ChebNet, GCN, MVGRL, GPRGNN and our MGCN on ACM and DBLP datasets via $t$-SNE algorithm \cite{T_SNE}. Here, we randomly select two categories of all samples so as to illustrate the margin of the corresponding decision boundaries clearly in Fig.\ref{t-sne}. From these results, we conclude that our proposed method has a larger margin of the decision boundaries compared with others. 
\vspace{-3pt}

\subsubsection{Visualization of Node Similarity Matrices}

We plot the heat maps of sample similarity matrices in the latent space to intuitively show the representation collapse problem in graph node classification methods and the effectiveness of our solution to this issue on DBLP and AMAP datasets. Here, we sort all samples by categories to make those from the same cluster beside each other. As illustrated in Fig. \ref{similarity}, we observe that GCN \cite{GCN} and GPRGNN \cite{GPRGNN} would suffer from representation collapse during the process of node encoding. Unlike them, our proposed method learns the more discriminative latent features, thus avoiding the representation collapse.

\section{Conclusion}

In this work, we propose a novel graph contrastive learning method termed {Interpolation-based Correlation Reduction Network (MGCN)} to alleviate the representation issue in semi-supervised node classification task. Specifically, we propose a graph interpolation module to force the prediction model to change linearly between samples, thus enlarging the margin of decision boundaries. Besides, the proposed correlation reduction module aims to keep the cross-view consistency of the embeddings. Benefited from these two modules, our network is guided to learn more discriminative representations, thus alleviating the representation collapse problem. Extensive experiments on six datasets demonstrate the superiority of our proposed methods. {In MGCN, the designed augmentation strategy interpolates the embeddings and associated labels. As a future work direction, it would be meaningful to explore how to adapt the interpolation-based augmentation strategy to unsupervised tasks.}

\section{Acknowledgments}

This work was supported by the National Key R\&D Program of China (no. 2022ZD0209103), the National Natural Science Foundation of China (no. 62325604, 62276271), the Postgraduate Scientific Research Innovation Project in Hunan Province (No. CX20220076).

\bibliographystyle{ACM-Reference-Format}
\bibliography{sample-base}

\end{document}